\documentclass[10pt,twocolumn,letterpaper]{article}

\usepackage[pagenumbers]{wacv}

\usepackage{graphicx}
\usepackage{amsmath}
\usepackage{amssymb}
\usepackage{booktabs}

\usepackage{caption}
\usepackage{subcaption}
\usepackage{tikz}
\usepackage{overpic}
\usepackage{bm}
\usepackage{multirow}

\makeatletter
\robustify\@latex@warning@no@line
\makeatother
\usepackage{authblk}

\usepackage[pagebackref,breaklinks,colorlinks]{hyperref}

\usepackage[capitalize]{cleveref}

\makeatletter
\renewcommand\AB@affilsepx{, \protect\Affilfont}
\makeatother

\newcommand\ver[1]{\rotatebox[origin=c]{90}{#1}}

\newcommand{\yes}{\checkmark}
\newcommand{\transpose}{^{\mathsf{T}}}

\newcommand{\vect}[1]{\bm{#1}}
\newcommand{\vecnorm}[1]{\left\|#1\right\|}

\newcommand{\PAR}[1]{\vskip2pt \noindent{\bf #1~}}

\newcommand{\no}{$\times$}

\crefname{section}{Sec.}{Secs.}
\Crefname{section}{Section}{Sections}
\Crefname{table}{Table}{Tables}
\crefname{table}{Tab.}{Tabs.}

\begin{document}

\title{Sun Off, Lights On:\\Photorealistic Monocular Nighttime Simulation for Robust Semantic Perception}

\author[1]{Konstantinos Tzevelekakis}
\author[2]{Shutong Zhang}
\author[1,3,4]{Luc Van Gool}
\author[1]{Christos Sakaridis}

\affil[1]{ETH Z\"urich}
\affil[2]{University of Toronto}
\affil[3]{KU Leuven}
\affil[4]{INSAIT}

\maketitle

\begin{abstract}
Nighttime scenes are hard to semantically perceive with learned models and annotate for humans. Thus, realistic synthetic nighttime data become all the more important for learning robust semantic perception at night, thanks to their accurate and cheap semantic annotations. However, existing data-driven or hand-crafted techniques for generating nighttime images from daytime counterparts suffer from poor realism. The reason is the complex interaction of highly spatially varying nighttime illumination, which differs drastically from its daytime counterpart, with objects of spatially varying materials in the scene, happening in 3D and being very hard to capture with such 2D approaches. The above 3D interaction and illumination shift have proven equally hard to \emph{model} in the literature, as opposed to other conditions such as fog or rain. Our method, named Sun Off, Lights On (SOLO), is the first to perform nighttime simulation on single images in a photorealistic fashion by operating in 3D. It first explicitly estimates the 3D geometry, the materials and the locations of light sources of the scene from the input daytime image and relights the scene by probabilistically instantiating light sources in a way that accounts for their semantics and then running standard ray tracing. Not only is the visual quality and photorealism of our nighttime images superior to competing approaches including diffusion models, but the former images are also proven more beneficial for semantic nighttime segmentation in day-to-night adaptation. Code and data will be made publicly available.
\end{abstract}

\section{Introduction}
\label{sec:intro}

A key requirement for level-5 automated driving systems and other outdoor autonomous agents is the robust visual perception of the surrounding scene, so that the content of the scene can be parsed under any visual condition~\cite{SAE:J3016}. However, the ubiquitous condition of night time has a detrimental effect on the quality of camera measurements due to effects such as underexposure, overexposure, and motion blur~\cite{zendel2017how}. This low input quality at night translates to a drastic deterioration in the performance of semantic perception algorithms for central tasks such as semantic segmentation, as compared to normal conditions or even other adverse conditions such as fog, rain, or snow~\cite{sakaridis2021acdc}. What makes things worse is the increased difficulty in the manual pixel-level semantic annotation of real nighttime images due to the above effects, which leads to errors and reduced image coverage in ground-truth annotations, in turn with negative impact on the reliability of models trained on such data.

As a result, a widely used paradigm for robust semantic nighttime segmentation is unsupervised domain adaptation (UDA) from day to night~\cite{sakaridis2019guided,sakaridis2022mapguided,wu2021dannet,xie2023sepico,hoyer2023mic,sakaridis2023ciss,gong2023continuous,sakaridis2021acdc,dai2018dark}. In this paradigm, both labeled -- thanks to easier acquisition and annotation -- daytime, or source-domain, images and unlabeled nighttime, or target-domain, images are available at training. A core element of such UDA methods is input-level adaptation~\cite{sankaranarayanan2018learning,li2019bidirectional,hoffman2018cycada} of source-domain images to the style of the target domain, so that the labels which are inherited by these adapted source-domain images can constrain the semantic segmentation model more effectively on closely-resembling real target-domain images. The three main approaches to such input-level adaptation are physically-based domain translation, learned image translation, and hand-crafted domain transformation.

On the one hand, learned, data-driven models for translating an input image, based e.g.\ on generative adversarial networks~\cite{zhu2017unpaired} or diffusion models~\cite{zhang2023adding}, can implicitly capture the statistics and patterns in the source and target domain and have proven very successful for input-level adaptation in synthetic-to-real UDA~\cite{hoffman2018cycada}. However, images adapted with such approaches are not photorealistic, as the latter do not model illumination, which changes drastically from day to night, or the properties of the scene, i.e.\ its 3D geometry and materials, which affect the spatially-varying interaction of light with the scene at night time and the resulting appearance of the image. The same limitation applies for hand-crafted methods for domain transformation, such as Fourier-based adaptation~\cite{yang2020fda}. On the other hand, while physically-based approaches for domain translation do not require training or reference-style images and have enjoyed remarkable success in condition-level UDA in the cases of fog~\cite{sakaridis2018semantic,bijelic2020seeing,hahner2021fog}, rain~\cite{halder2019physics}, and snowfall~\cite{hahner2022lidar}, no such approach has been proposed for the ubiquitous nighttime condition to the best of our knowledge.

\begin{figure*}[tb]
    \centering
    \includegraphics[clip, trim=1.7cm 18.4cm 1.7cm 0.2cm, width=0.88\textwidth]{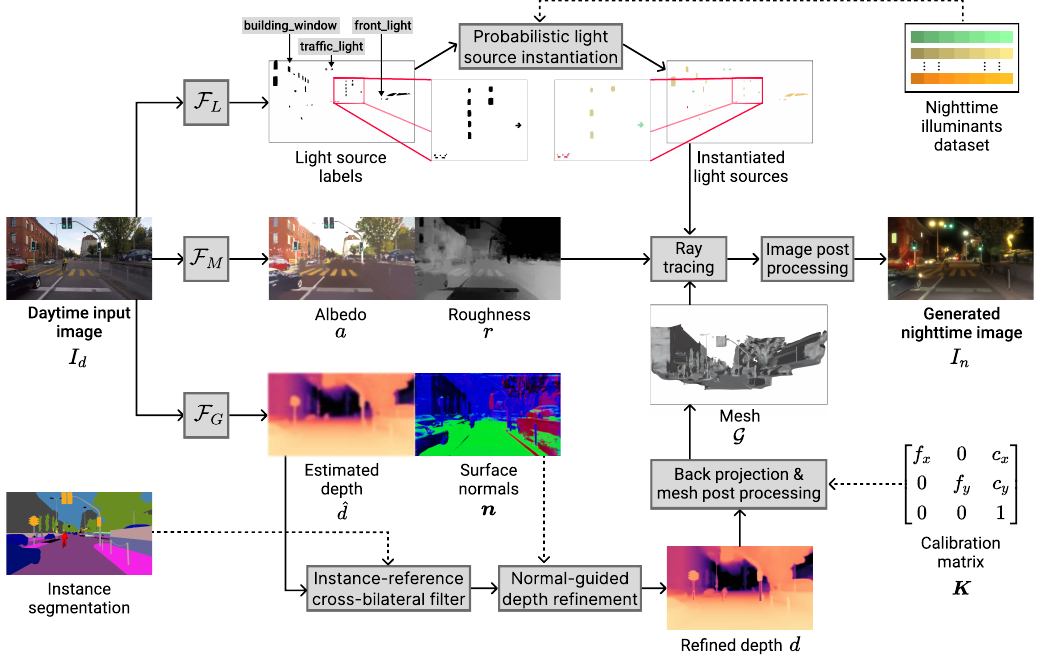}
    \vspace{-0.2cm}
    \caption{\textbf{Overview of SOLO.} Our method accepts as input a single daytime image $I_d$. Geometric $(\hat{d}, \vect{n})$ and material $(a, r)$ representations are estimated with the inverse rendering networks $\mathcal{F}_G$ and $\mathcal{F}_M$, respectively. A light source segmentation network $\mathcal{F}_L$ predicts the regions in $I_d$ which correspond to inactive light sources that may emit light at night. The initial depth map $\hat{d}$ is filtered and optimized with guidance from an instance semantic segmentation mask and the estimated surface normal map $\vect{n}$, respectively. The refined depth map $d$ and the camera intrinsics are used to construct the 3D scene mesh $\mathcal{G}$. Nighttime light sources in the scene are instantiated probabilistically group-wise, using the predictions of $\mathcal{F}_L$ to sample their activation variables and the external real-world nighttime illuminants dataset we have collected to set their chromaticities. The activated light sources, the materials $(a, r)$ and the 3D mesh $\mathcal{G}$ are finally fed to the ray tracing module which renders a raw image that is subsequently post-processed in a standard fashion to compute the output nighttime image $I_n$.}
    \label{fig:overview}
    \vspace{-0.5cm}
\end{figure*}

In this paper, we present the first physically-based monocular approach to nighttime simulation on real daytime images, aiming to afford photorealistic synthetic nighttime counterparts. We pursue this through inverse rendering, probabilistic light source instantiation, and physically-based rendering via ray tracing. Our rationale for photorealistic nighttime simulation is to (i) estimate the scene representations required for ray tracing from the input daytime image, notably the positions of all inactive light sources in the scene, (ii) modify the lighting of the scene by removing the sun and probabilistically activating the aforementioned light sources in a semantics-aware fashion (hence the name of our method Sun Off, Lights On or SOLO), and (iii) relight the scene by running ray tracing with the updated, nighttime lighting to render the nighttime image.

The key novel contributions of SOLO are (i) the proposed semantics-aware probabilistic light source instantiation for shifting the lighting of the scene from day time to night time, (ii) a carefully crafted, normals- and semantics-guided optimization for depth map refinement within the mesh-based 3D reconstruction of our monocular inverse rendering module, as well as (iii) our overall physically-based monocular nighttime simulation pipeline which elegantly combines inverse rendering and ray-tracing-based relighting.
Our synthetic nighttime images match the appearance of real nighttime images better thanks to their photorealism and thus serve as a good proxy for the real nighttime domain.
We verify this superiority of SOLO in an objective fashion, by using it as the input-level adaptation module of a state-of-the-art UDA pipeline~\cite{hoyer2022hrda} for day-to-night semantic segmentation adaptation on the challenging ACDC-Reference$\to$ACDC-night~\cite{sakaridis2024acdc} benchmark.

\section{Related Work}

\PAR{Scene relighting}\,is a fundamental task in computer vision and graphics. In the context of \emph{novel view synthesis}, it refers to rendering a scene from different camera views~\cite{gao2023relightable, srinivasan2021nerv, lyu2022neural}. Another version of scene relighting involves rendering a scene for another time of day or under varying but known lighting conditions \cite{wang2023neural, srinivasan2021nerv, zhang2022simbar}. Our setting is a special case of the latter, performing photorealistic nighttime simulation by considering ambient lighting, light sources activated at night, and their interactions with the scene.

\PAR{Relighting through inverse rendering.}\,Conventionally, relighting a scene requires accurate estimation of geometry and material parameters, a process known as inverse rendering, a classic underconstrained problem in computer vision. The most prevalent approach is to learn priors on the shape, illumination and reflectance, using large labeled image datasets for geometry and materials like in \cite{zhang2022simbar, li2020inverse,  sang2020single, wei2020object, sengupta2019neural, yu2019inverserendernet}. Scene relighting is then achieved by forward rendering. SOLO is closely related to these methods by treating state-of-the-art inverse rendering models as black boxes to estimate both material and geometry parameters. Unlike the aforementioned works, SOLO also employs a semantic light source segmentation model, enabling semantically aware explicit reasoning on both the activation and color properties of the light sources. Since the emergence of \emph{neural radiance fields} (NeRFs), a new approach to scene relighting has gained traction. Although the seminal work on NeRFs \cite{mildenhall2021nerf} did not handle relighting, recent works \cite{gao2023relightable, wang2023neural, lyu2022neural, srinivasan2021nerv} have reformulated the continuous volumetric function to accommodate it. An additional feature of NeRF-based approaches is the ability to recover full 3D models using a sparse set of multi-view images as input. However, in our setting, only one image per scene is available, making NeRF-based approach not easily applicable.

\PAR{Day-to-night transfer in 2D.}\,A large body of literature, simulates nighttime by employing generative models for \emph{style transfer}. These models are either based on the generative adversarial network (GAN) architecture \cite{zhu2017unpaired, huang2018auggan, cherian2019sem}, or on the more recent diffusion architecture \cite{zhang2023adding}. However, during the day-to-night translation, given only a 2D daytime image, these purely data-driven models struggle to account for the activation of light sources, the 3D interactions of light rays with objects in the scene, the rendering of spatially varying illumination, and the simulation of under- or over-exposure. Therefore, even recent diffusion-based architectures \cite{zhang2023adding} cannot accurately simulate nighttime conditions, as evidenced in our experiments. Notable \emph{hand-crafted} 2D-based approaches also exist. In \cite{punnappurath2022day}, a framework processes a given daytime image by introducing artificial light sources sampled from a nighttime illuminants dataset. Additionally, in \cite{yang2020fda}, a method for UDA is presented. This method, based on the Fourier Transform, reduces the shift in appearance from the source to the target image by swapping the low-frequency components of the source magnitude spectrum with those from the target magnitude spectrum. This allows the source images to adopt the global appearance characteristics (e.g., texture, lighting conditions). However, both methods strictly operate in the 2D space and fail to provide photorealistic results.

\section{Sun Off, Lights On}

The proposed nighttime simulation method, named ``Sun Off, Lights On'' (SOLO), is based on a \emph{single} daytime input image as shown in Fig.~\ref{fig:overview}. SOLO estimates the geometry and materials of the scene through inverse rendering (Sec.~\ref{sec:method:inverse_rendering}). The novel probabilistic light source instantiation module of SOLO (Sec.~\ref{sec:method:light_source_instantiation}) determines the lighting configuration of the nighttime scene. Light sources of different categories are first semantically segmented and grouped and then probabilistically activated to simulate nighttime lighting. Finally, the forward rendering module (Sec.~\ref{sec:method:forward_rendering}) uses the estimated geometry, materials, and lighting to perform physically-based rendering (PBR) and thus deliver a photorealistic nighttime image of the scene.

\subsection{Inverse Rendering}
\label{sec:method:inverse_rendering}

\subsubsection{Geometry and Materials Estimation}
\label{sec:method:inverse_rendering:material_geometry}

In this work, we utilize state-of-the-art off-the-shelf monocular estimation networks for both geometry and materials. More specifically, we consider depth $\hat{d}$ and surface normals $\hat{\vect{n}}$ maps as the dense geometric representations of the scene which are estimated by the geometry network as $F_G(I_d)=(\hat{d},\hat{\vect{n}})$. As per the material representations, these are diffuse albedo $a$ and specular roughness $r$, they are estimated by the material model as $F_M(I_d)=(a, r)$.

As SOLO is applied to daytime \emph{outdoor} scenes, it requires inverse rendering networks trained on such scenes. To the best of our knowledge, no real-world outdoor dataset with dense annotations for materials, in particular for roughness $r$, exists to perform the above training. However, material properties, such as albedo and roughness, are not a priori correlated with their occurrence in an indoor or outdoor scene. Thus, the implicit assumption made in the general form of our method is that materials output by an indoor-trained network are accurate for outdoor scenes as well. On the other hand, there is a large collection of geometric models that are trained on real-world outdoor sets. An important requirement stemming from ray tracing and PBR is the metric character of the reconstructed scene, so the depth units must be known. Thus, only network architectures which output metric depth maps are relevant for SOLO.
Since it is typical for the aforementioned models to output maps of lower resolution than the original daytime input image, upsampling is required both for geometry and material parameters. Standard bilinear interpolation is used for the geometric maps and a more sophisticated, joint bilateral upsampling \cite{joint:bilateral:upsampling} for the material maps. The latter utilizes the corresponding daytime image as reference and we found it performs better than plain bilinear interpolation.

\subsubsection{Depth Refinement}
\label{sec:method:inverse_rendering:depth_processing}
\PAR{Instance-Reference Cross-bilateral Filter.}\,Even a slight misalignment between an actual object boundary and the corresponding depth edge in the prediction $\hat{d}$ of the geometry network $F_G$ deteriorates the realism of the subsequent 3D reconstruction. To eliminate such misalignments, we adapt the dual-reference cross-bilateral filter of~\cite{sakaridis2018model}. Apart from spatial information, this filter originally exploits both a color and a semantic reference signal to refine a transmittance map akin to depth. For our depth filtering case, local variations in the color in $I_d$ do not necessarily correspond to variations in depth values. We thus drop the color reference of~\cite{sakaridis2018model} and only use its semantic reference. In particular, we replace the semantic reference labels in~\cite{sakaridis2018model} with instance-level semantic reference labels, in order to properly preserve depth edges between different objects of the same semantic class which are adjacent to each other.
To formulate our instance-reference cross-bilateral filter, we use $\vect{p}$ to denote any non-boundary pixel in the depth map $\hat{d}$, and $\vect{q}$ to denote a pixel belonging to the neighborhood $\mathcal{N}$ of $\vect{p}$. The filtered depth $\Tilde{d}$ at a pixel location $\vect{p}$ is computed as a weighted average of initial depth values $\hat{d}$:
\begin{equation}
\label{eqn:cross_bilateral_filter}
\Tilde{d}(\vect{p}) = 
                \frac{ 
                    \sum_{\vect{q} \in \mathcal{N}(\vect{p}) }
                        G_{\sigma_s}(\|\vect{q} - \vect{p}\|) \delta(h(\vect{q}) - h(\vect{p}))
                    \hat{d}(\vect{q})
                }{
                    \sum_{\vect{q} \in \mathcal{N}(\vect{p}) }
                    G_{\sigma_s}(\|\vect{q} - \vect{p}\|) \delta(h(\vect{q}) - h(\vect{p}))
                } ,
\end{equation}
where $G_{\sigma_s}$ is the spatial Gaussian kernel applied on the $l_2$ distance between pixels. This is done only when the instance labels $h$ of the corresponding pixels match, as dictated by the Kronecker delta term $\delta$.

\PAR{Uncertain Depth Regions.}\,For the depth and mesh refinement steps, we introduce the concept of \emph{uncertain} depth regions $U$. These are regions located near object boundaries, where accurate depth prediction is challenging. Boundaries of thin objects such as people and traffic signs, are typical associated examples. To locate those regions, a sliding window approach is employed, with a $k\times k$ window. All pixels contained in a window are marked as \emph{uncertain} only if both of the following criteria are satisfied: (i) at least two semantic segments overlap with the window, and (ii) the variance of depth values across the window is larger than a predefined threshold $t$. Finally, objects that are located further from the camera than a distance threshold $r$ are disregarded.

\PAR{Surface-Normal-Guided Depth Optimization.}\,The surface normals $\vect{n}$ output by the geometry network $F_G$ provide additional fine-grained geometric information for accurate mesh-based 3D reconstruction of the input scene, on top of depth. In this work, we devise a novel optimization method which exploits this information from normals to refine the depth map. To define the optimization objective, we first model how surface normals $\vect{n}$ can be inferred from a corresponding depth map $z(x, y)$. We assume a standard pinhole camera model
where $u = \frac{f_x}{z}x+c_x$ and $v = \frac{f_y}{z}y+c_y$ are pixel-space coordinates, $x$ and $y$ are 3D camera-frame coordinates, $z$ is the depth at $(x, y)$, $f_x$ and $f_y$ are the focal lengths, and $c_x$ and $c_y$ are the principal point coordinates.
The surface normal vector $\vect{n}$ is perpendicular to the tangent plane of the 3D surface at $(x,y,z(x,y))$. To obtain the direction vector $\vect{s}$ of this plane, we use the graph function $\vect{\mathcal{F}}\left(x, y\right) = \left(x, y, z(x, y)\right)$ of $z$, and its gradient $\nabla\vect{\mathcal{F}}\left(x, y\right) = {(\frac{\partial \vect{\mathcal{F}}}{\partial x},\,\frac{\partial \vect{\mathcal{F}}}{\partial y})}\transpose$.
The direction vector $\vect{s}$ of the tangent plane is perpendicular to both rows of $\nabla\vect{\mathcal{F}}$, so
\begin{equation}
\vect{s} = \frac{\partial \vect{\mathcal{F}}}{\partial x} \times \frac{\partial \vect{\mathcal{F}}}{\partial y} = (-\frac{\partial z}{\partial x}, -\frac{\partial z}{\partial y}, 1).
\end{equation}
Finally, the surface normal $\vect{n}$ is obtained by normalizing $\vect{s}$.
Our optimization loss $\mathcal{L}$ is defined as a weighted sum of two terms. The first term, $L_1$, is formulated as:
\begin{equation}
\label{eqn:L_1}
L_1 = 
\frac{1}{m}\sum_{\vect{p}} {\vecnorm{\nabla\vect{\mathcal{F}}(\vect{p})\hat{\vect{n}}(\vect{p})}}_2^2 \bar{U}(\vect{p}),
\end{equation}
where $\hat{\vect{n}}$ are the surface normals initially predicted by $F_G$, $\bar{U}$ is the complement of the binary \emph{uncertain} depth region mask, and $m$ is the total number of pixels.
The role of $\bar{U}$ is to ignore depth discontinuities in $L_1$. Minimizing $L_1$ modifies the depth map $\tilde{d}$ to better conform to the independently predicted normals $\hat{\vect{n}}$, yielding a more faithful 3D mesh.
To avoid smoothing out salient depth details completely and to balance the effect that potential inaccuracies in predicted normals $\hat{n}$ have, the second term of our optimized loss, $L_2$, quantifies the error between the depth map $z$ which is under optimization and the filtered depth map $\tilde{d}$ as:
\begin{equation}
\label{eqn:L_2}
L_2 = \frac{1}{m}\sum_{\vect{p}} \left(\tilde{d}(\vect{p}) - z(\vect{p})\right)^2.
\end{equation}
The complete optimization loss is $\mathcal{L} = \lambda_1 L_1 + \lambda_2 L_2$,
where $\lambda_1$ and $\lambda_2$ are the weights for the corresponding loss terms. Note that $z$ is initialized with the filtered depth map $\tilde{d}$. We denote the final, optimized depth map by $d$.

\subsubsection{Backprojection and Mesh Post-processing}
\label{sec:method:inversre_rendering:back_proj_and_mesh_processing}

To initialize the 3D mesh $\mathcal{G}$ to be used subsequently for ray tracing, we use the backprojection equation $\vect{x} = d  \vect{K}^{-1} \bar{\vect{p}}$, where $\bar{\bm{p}} = \left(u, v, 1 \right)\transpose$ denotes the homogeneous representation of the pixel coordinates, $d$ is the final depth map, $\vect{x}$ is the 3D scene point serving as a vertex of $\mathcal{G}$, and $\vect{K}$ is the calibration matrix.
Although this initial mesh is faithful to the geometry of the 3D scene from the camera's point of view, the monocular information it captures results in geometric errors with spurious faces for occluded objects or objects outside the field of view. Such spurious faces obstruct light rays, creating erroneous shadows at ray tracing. To mitigate this, we apply additional post-processing to remove these faces and subsequently restore a watertight mesh.

\subsection{Probabilistic Light Source Instantiation}
\label{sec:method:light_source_instantiation}

To illuminate a 3D scene photorealistically, semantic information specific to its light sources is required to assign light source attributes via stochastic rules.
Additionally, we construct a dataset of nighttime illuminants
to provide realistic chromaticities. Light source strengths are sampled from empirically defined intervals.

\PAR{Light source segmentation dataset and model.}\,To the best of our knowledge, no outdoor dataset with light source annotations exists. We first define a novel, comprehensive light source taxonomy for outdoor scenes building upon the object taxonomy of Cityscapes~\cite{cordts2016cityscapes}, with 16 light source categories. We then annotate a reasonably-sized daytime set with pixel-level light source labels for this taxonomy by segmenting active \emph{and} inactive light sources in its images. We finally fine-tune a normal semantic segmentation model on this labeled set, using a new prediction layer to account for the different taxonomy. The resulting light source segmentation model $\mathcal{F}_L$ predicts light source labels for the complete daytime set on which we apply SOLO.

\PAR{Nighttime illuminants dataset.}\,The color appearance of a nighttime illuminant can be specified in terms of the $xyY$ color space. 
Our nighttime illuminants dataset $\mathcal{N}$ consists of real-world chromaticity samples for each light source category, collected using a gray card and a DSLR camera, following~\cite{punnappurath2022day}. Each sample includes a raw image of the gray card, illuminated by an instance of the sampled light source category. To avoid pollution from neighboring sources, we made sure that only the light source to be sampled was visible from the surface of the gray card during collection. The captured raw images are processed with a standard camera pipeline to obtain chromaticity~\cite{punnappurath2022day, rowlands2020color, sumner2014processing}.

\PAR{Light source instantiation module.}\,Our probabilistic instantiation module assigns attributes to light sources of the scene based on stochastic rules, conditioned by semantic and instance information. This information is incorporated in two ways: (i) by leveraging the light source label, such as ``vehicle front light'', and (ii) by exploiting the instance-level semantic label, such as ``car 2''. By combining these attributes, a tree structure is constructed, which specifies the light source group that a light source belongs to. More specifically, each node of this tree can either correspond to (i) a light source group or (ii) a light source/leaf node, whereas the edges of the tree indicate membership. For example, two ``vehicle front lights'' can both be children nodes of the same ``car'' light source group. Three attributes are assigned to each light source: the chromaticity, the strength, and the probability of activation $y$. These attributes primarily depend on the light source category. Moreover, it is plausible for light sources belonging to the same group, e.g.\ the front left and front right lights of a car, to share the same attributes. Light source attributes are modeled as random variables. In particular, chromaticity follows a discrete uniform distribution over the relevant samples of $\mathcal{N}$, whereas probability of activation and strength follow continuous uniform distributions over empirically defined intervals. In particular, the stochastically sampled probability of activation $y$ is in turn used to define another Bernoulli variable that models the actual activation of the light source. That is, the random activation variable $X$ for a light source follows $X\sim\text{Bernoulli}\left(y\right)$,
where the Bernoulli parameter $y$ is the realization of the intermediate variable $Y\sim\text{Uniform}\left(a, b\right)$.

\subsection{Forward Rendering}
\label{sec:method:forward_rendering}

All constituents of the scene are combined via forward rendering to derive a photorealistic nighttime image. The input is the mesh $\mathcal{G}$ overlayed with the estimated materials and active light sources, as shown in Fig.~\ref{fig:overview}. Moreover, head lights from the ego-vehicle are simulated in order to enhance realism. We run standard ray tracing to generate a linear image. The latter is fed to a standard post-processing pipeline to yield the photorealistic nighttime image.

\PAR{Physically-based rendering and ray tracing}\,is formulated with a specialized version of the rendering equation, i.e.\ the reflectance equation:
\begin{equation}
\begin{aligned}
L_o(\vect{x}, \vect{\omega}_o) &= L_e(\vect{x}, \vect{\omega}_o) + L_r(\vect{x}, \vect{\omega}_o), \\
L_r(\vect{x}, \vect{\omega}_o) &= \int_{\Omega} f_r(\vect{x}, \vect{\omega_o}, \vect{\omega_i}) L_i(\vect{x}, \vect{\omega}_i) (\vect{\omega}_i \cdot \vect{n})d\vect{\omega}_i,
\end{aligned}
\end{equation}
where for a point $\vect{x}$ on a surface, $L_o$, $L_e$, and $L_r$ denote the outgoing, emitted, and reflected radiance, respectively. Moreover, $\vect{\omega}_o$ and $\vect{\omega}_i$ correspond to outgoing and incident light directions respectively, $\vect{n}$ is the normal vector, and $f_r$ is the bidirectional reflectance distribution function (BRDF).
In particular, we employ the physically motivated ``Disney'' BRDF proposed in~\cite{burley2012physically} and later adopted by Unreal Engine 4~\cite{karis2013real}.
This BRDF is formulated as
\begin{equation}
\label{eqn:brdf}
f_r(\vect{x}, \vect{\omega_i}, \vect{\omega_o}) = f_d(\vect{x}, \vect{\omega_i}, \vect{\omega_o}) + f_s(\vect{x}, \vect{\omega_i}, \vect{\omega_o}),
\end{equation}
where $f_d$ and $f_s$ are the diffuse and specular BRDF components.
For ray tracing, we also model the directionality of the light sources, so that both strongly directional and rather diffuse light sources can be simulated. To this end, the strength of a light source is weighted by a function $g$ of the incident direction $\vect{v}$ and of the normal vector $\vect{\hat{n}}$ of the \emph{area light source} surface, formulated as $g\left(\vect{v}, \vect{\hat{n}}\right) = \cos\left(\pi\left(| \vect{v} \cdot \vect{\hat{n}} | - 1\right)/2\right)$. It follows that when the outgoing ray is aligned with the surface normal, the outgoing radiance is maximal, whereas when the associated angle approaches $\pm \frac{\pi}{2}$, the outgoing radiance goes to 0.

\PAR{Image post-processing.}\,After acquiring a linear-color, noise-free image from ray tracing, we employ a post-processing pipeline imitating the steps of a standard image signal processor (ISP). This serves two purposes. On the one hand, the appropriate transformations should be applied to make the generated image displayable. On the other hand, since our nighttime images are meant for inputs of neural networks, the visual artifacts present in real nighttime images should also appear in our simulated images to minimize the distribution shift. Our post-processing pipeline starts by adjusting the brightness of the image, setting the exposure appropriately. Since the image is in the linear XYZ color space, the standard Bradford color adaptation method~\cite{lam1985metamerism} along with gamma correction are used to transform the image to the sRGB color space. Moreover, fog glare is incorporated to make the appearance of light sources more realistic and noise is added using the standard heteroscedastic Gaussian noise model, following~\cite{punnappurath2022day}.

\section{Experiments}
\label{sec:exp}

\subsection{Implementation Details}

The state-of-the-art pre-trained model of~\cite{li2020inverse} is employed as the materials estimation network $F_M$. However, this model has limitations. First, its indoor training data do not include several materials commonly found in outdoor scenes. Second, these training data only include dielectric materials, leading to low-quality material maps, especially in regions with \emph{metallic} objects. We experimented with more recent indoor trained models, such as that of~\cite{zhu2022learning}, but they also estimated materials poorly. We attribute these shortcomings to the large distribution shift between the materials in indoor training sets and those in real-world outdoor scenes. As a result, the roughness estimates are not sufficiently accurate. Notably, the specular microfacet term $f_s$ in \eqref{eqn:brdf} is very sensitive to the roughness value. Thus, in all our experiments, we revert to using only the diffuse BRDF component $f_d$ in \eqref{eqn:brdf}. 
We use the state-of-the-art pre-trained UniDepth~\cite{piccinelli2024unidepth} and iDisc~\cite{piccinelli2023idisc} networks to predict depth and surface normals, respectively. For depth refinement, $\sigma_s$ is set to 5px in \eqref{eqn:cross_bilateral_filter}. For uncertain depth regions, we set $k$=10px, $t$=0.01, and $r$ to the mid-range of the scene's depth. Normal-guided depth optimization uses Adam~\cite{kingma2014adam} for 1000 iterations with a learning rate of 2e-4. We set $\lambda_1$=50 and $\lambda_2$=1 in this optimization. The camera intrinsics are $f_x$=$f_y$=1780px, $c_x$=959.5px and $c_y$=539.5px.
For PBR via ray tracing, we adopt the multi-scattering GGX implementation~\cite{heitz2016multiple} of the Cycles path tracer~\cite{cycles}, providing off-the-shelf physically based results.
In the post-processing pipeline, we set exposure to 3.25 stops, gamma to 2.2, and employ the implementation of~\cite{punnappurath2022day} for noise addition.

\begin{figure*}[tb]
    \centering
    \begin{subfigure}[b]{0.193\textwidth}
        \begin{overpic}[width=\textwidth]{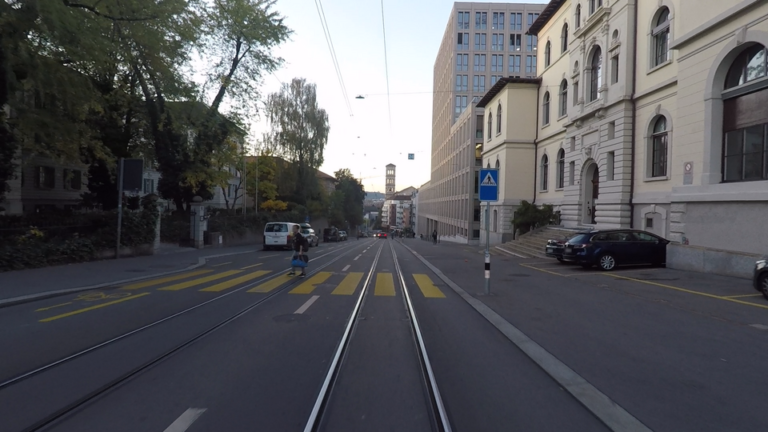}
            \put(6, 48){
                \begin{tikzpicture}[overlay]
                \node[circle, fill=white, inner sep=1pt, font=\scriptsize] (1) at (0,0) {\textbf{1}};
            \end{tikzpicture}
            }
        \end{overpic}
    \end{subfigure}
    \hfill
    \begin{subfigure}[b]{0.193\textwidth}
        \includegraphics[width=\textwidth]{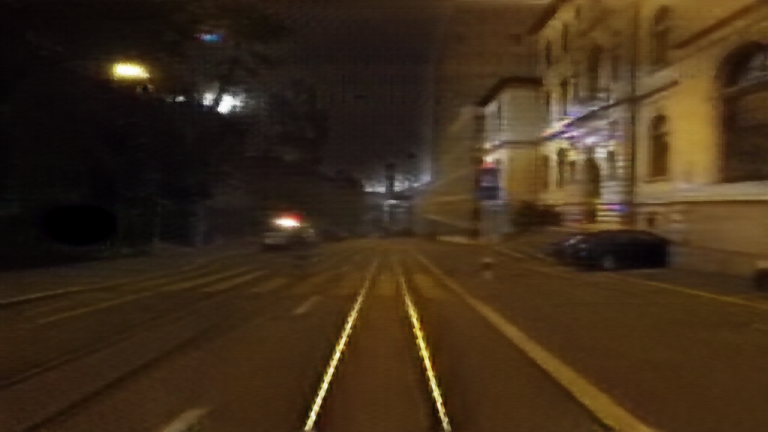}
    \end{subfigure}
    \hfill
    \begin{subfigure}[b]{0.193\textwidth}
        \includegraphics[width=\textwidth]{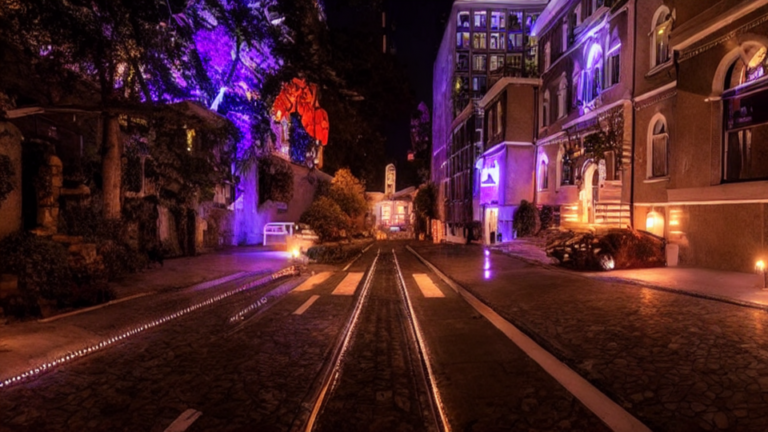}
    \end{subfigure}
    \hfill
    \begin{subfigure}[b]{0.193\textwidth}
        \includegraphics[width=\textwidth]{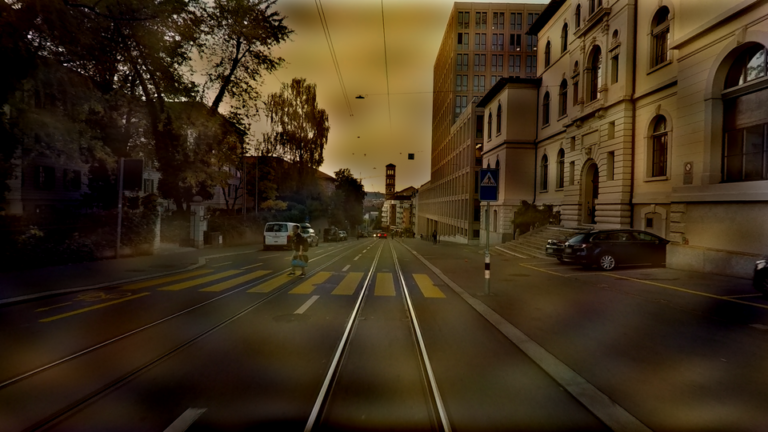}
    \end{subfigure}
    \hfill
    \begin{subfigure}[b]{0.193\textwidth}
        \includegraphics[width=\textwidth]{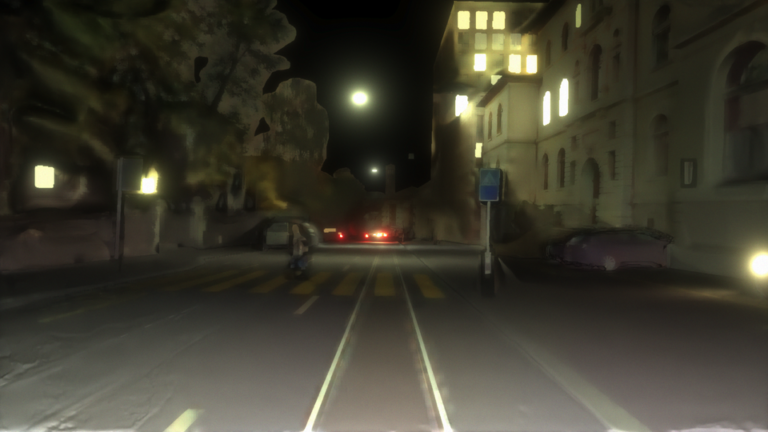}
    \end{subfigure}
    
    \vspace{0.1em}
                 
    \begin{subfigure}[b]{0.193\textwidth}
        \begin{overpic}[width=\textwidth]{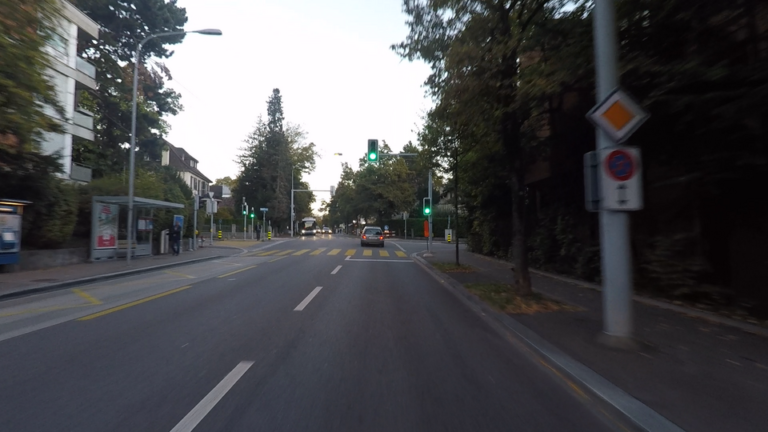}
            \put(6, 48){
                \begin{tikzpicture}[overlay]
                \node[circle, fill=white, inner sep=1pt, font=\scriptsize] (1) at (0,0) {\textbf{2}};
            \end{tikzpicture}
            }
        \end{overpic}
    \end{subfigure}
    \hfill
    \begin{subfigure}[b]{0.193\textwidth}
        \includegraphics[width=\textwidth]{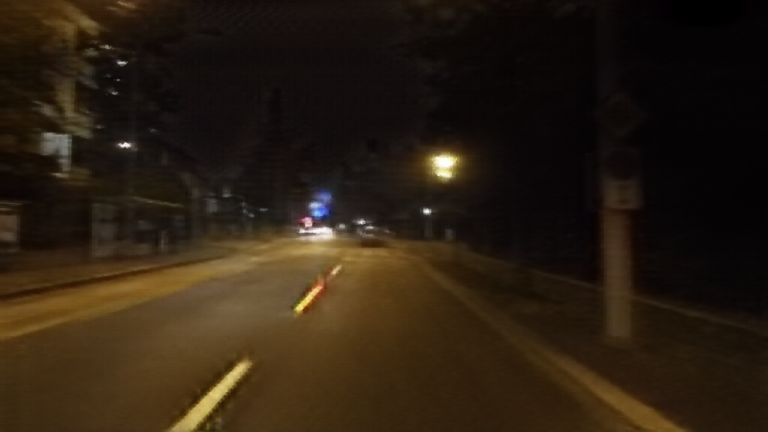}
    \end{subfigure}
    \hfill
    \begin{subfigure}[b]{0.193\textwidth}
        \includegraphics[width=\textwidth]{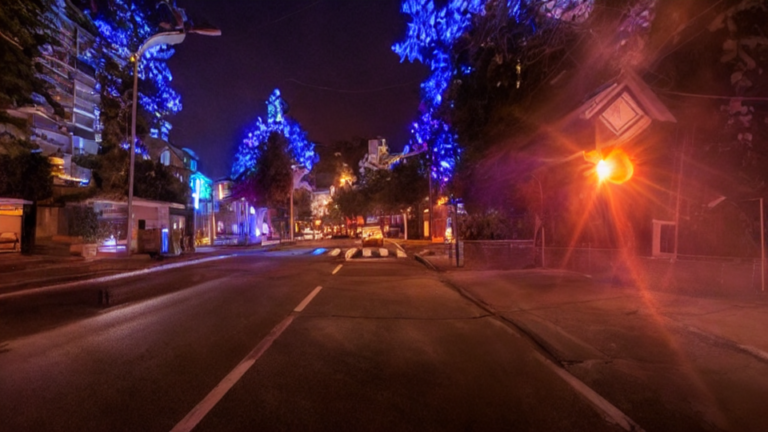}
    \end{subfigure}
    \hfill
    \begin{subfigure}[b]{0.193\textwidth}
        \includegraphics[width=\textwidth]{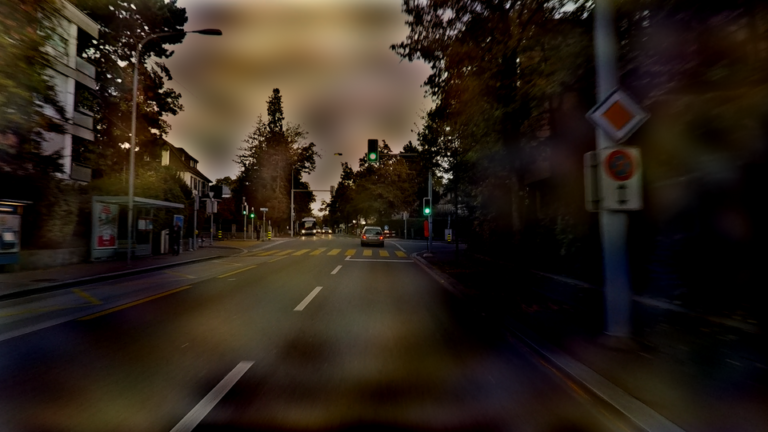}
    \end{subfigure}
    \hfill
    \begin{subfigure}[b]{0.193\textwidth}
        \includegraphics[width=\textwidth]{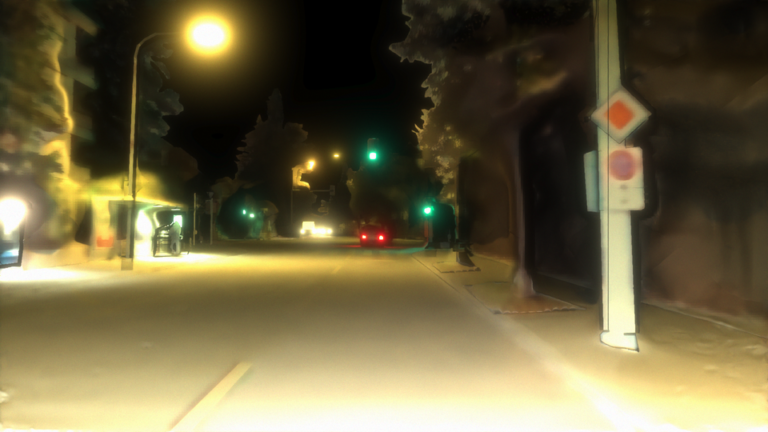}
    \end{subfigure}
    
    \vspace{0.1em}
        
    \begin{subfigure}[b]{0.193\textwidth}
        \begin{overpic}[width=\textwidth]{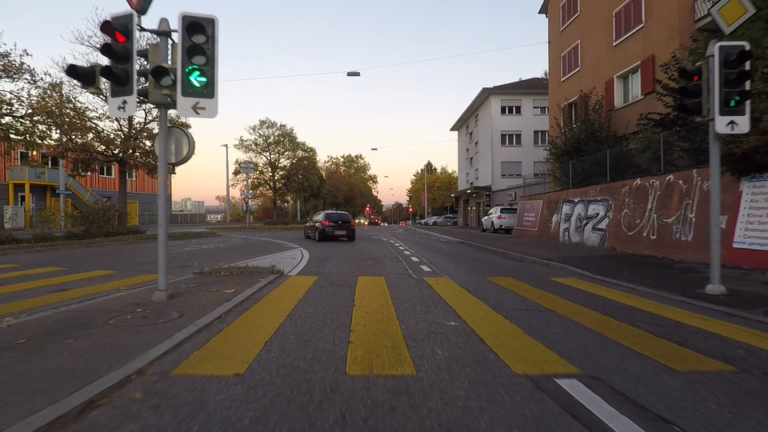}
            \put(6, 48){
                \begin{tikzpicture}[overlay]
                \node[circle, fill=white, inner sep=1pt, font=\scriptsize] (1) at (0,0) {\textbf{3}};
            \end{tikzpicture}
            }
        \end{overpic}
    \end{subfigure}
    \hfill
    \begin{subfigure}[b]{0.193\textwidth}
        \includegraphics[width=\textwidth]{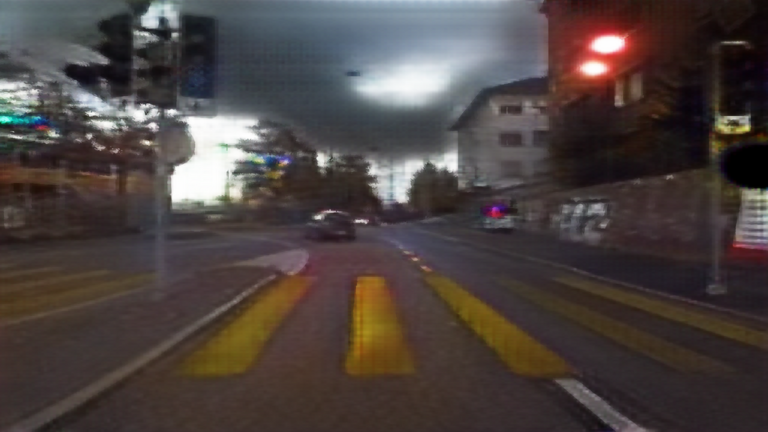}
    \end{subfigure}
    \hfill
    \begin{subfigure}[b]{0.193\textwidth}
        \includegraphics[width=\textwidth]{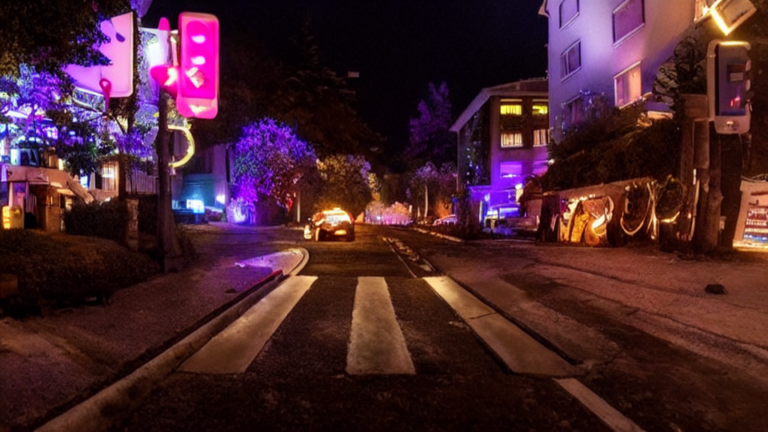}
    \end{subfigure}
    \hfill
    \begin{subfigure}[b]{0.193\textwidth}
        \includegraphics[width=\textwidth]{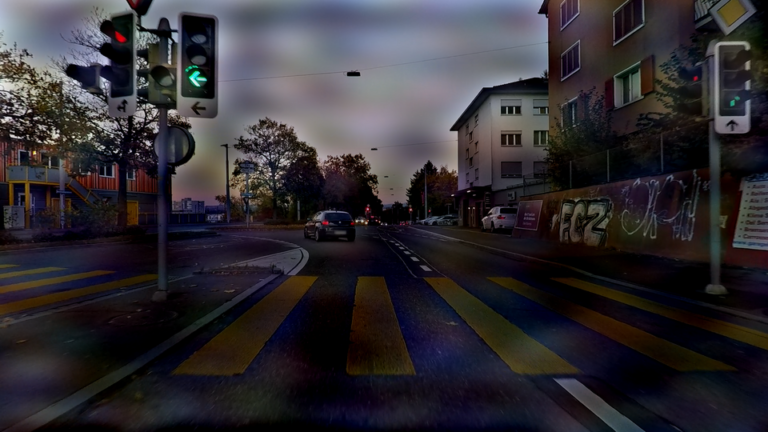}
    \end{subfigure}
    \hfill
    \begin{subfigure}[b]{0.193\textwidth}
        \includegraphics[width=\textwidth]{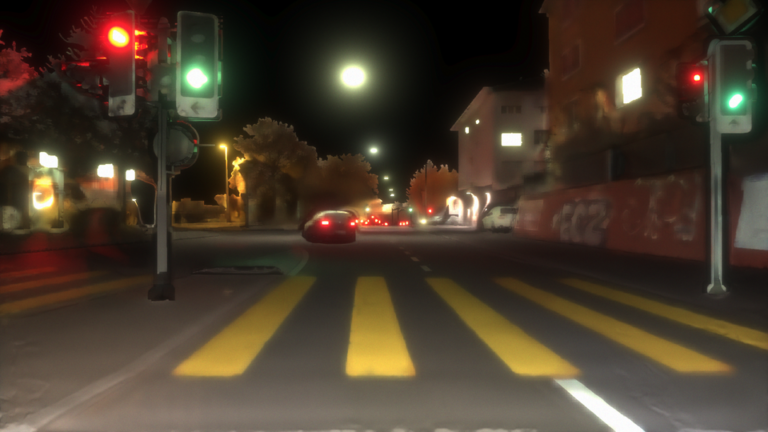}
    \end{subfigure}
    
    \vspace{0.1em}
        
    \begin{subfigure}[b]{0.193\textwidth}
        \begin{overpic}[width=\textwidth]{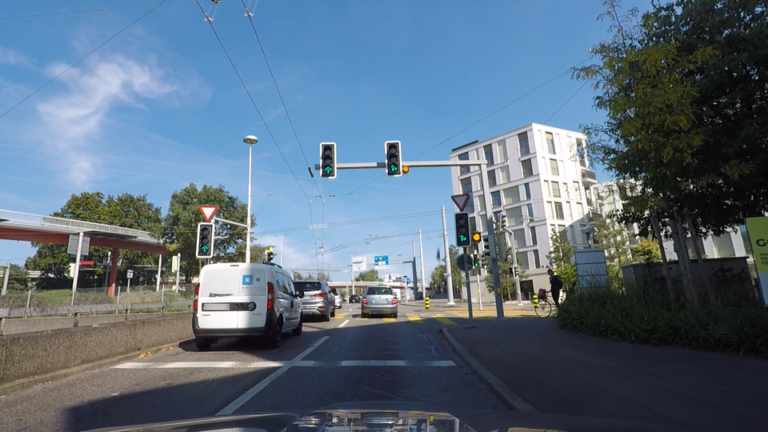}
            \put(6, 48){
                \begin{tikzpicture}[overlay]
                \node[circle, fill=white, inner sep=1pt, font=\scriptsize] (1) at (0,0) {\textbf{4}};
            \end{tikzpicture}
            }
        \end{overpic}
    \end{subfigure}
    \hfill
    \begin{subfigure}[b]{0.193\textwidth}
        \includegraphics[width=\textwidth]{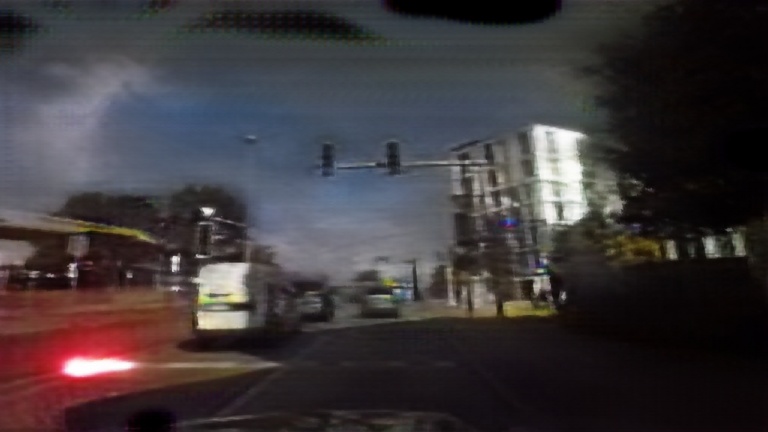}
    \end{subfigure}
    \hfill
    \begin{subfigure}[b]{0.193\textwidth}
        \includegraphics[width=\textwidth]{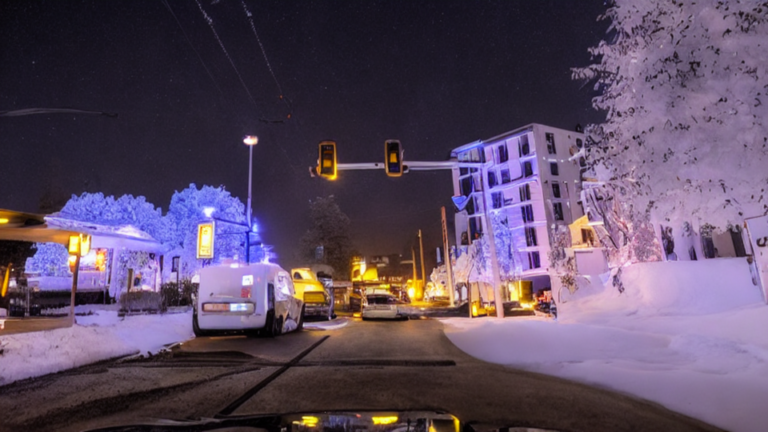}
    \end{subfigure}
    \hfill
    \begin{subfigure}[b]{0.193\textwidth}
        \includegraphics[width=\textwidth]{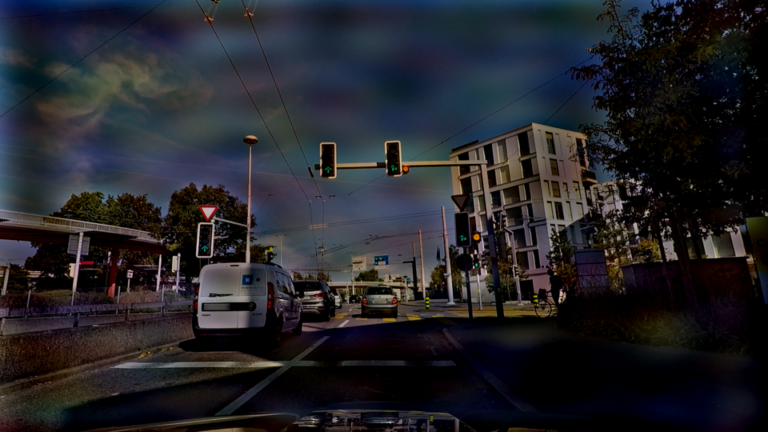}
    \end{subfigure}
    \hfill
    \begin{subfigure}[b]{0.193\textwidth}
        \includegraphics[width=\textwidth]{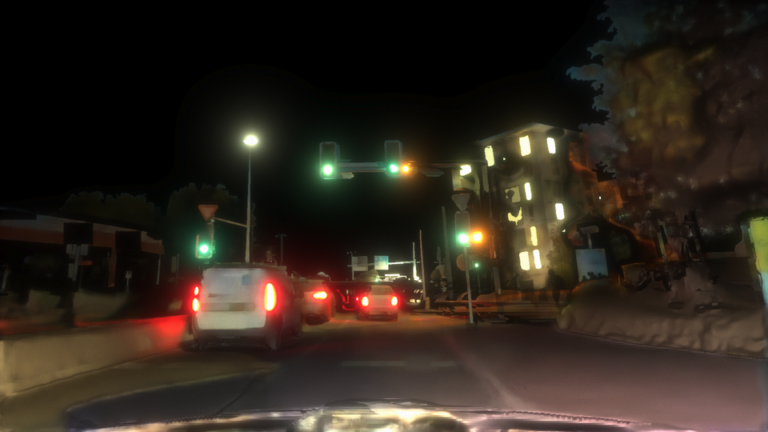}
    \end{subfigure}
     
     \vspace{0.1em}
    \begin{subfigure}[b]{0.193\textwidth}
        \begin{overpic}[width=\textwidth]{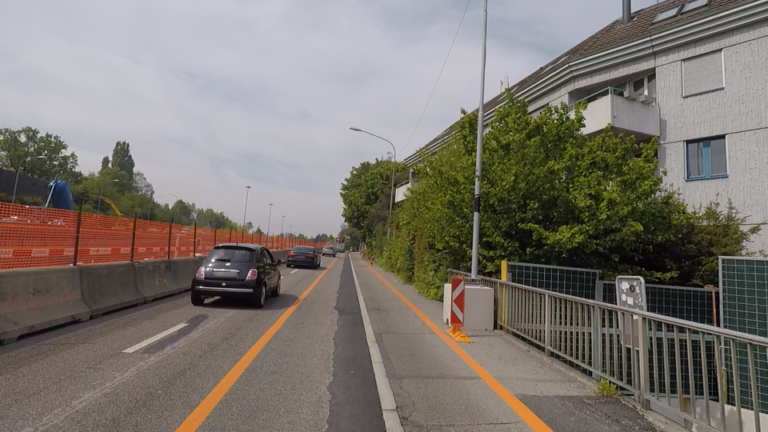}
            \put(6, 48){
                \begin{tikzpicture}[overlay]
                \node[circle, fill=white, inner sep=1pt, font=\scriptsize] (1) at (0,0) {\textbf{5}};
            \end{tikzpicture}
            }
        \end{overpic}
        \caption{Input image}
    \end{subfigure}
    \hfill
    \begin{subfigure}[b]{0.193\textwidth}
        \includegraphics[width=\textwidth]{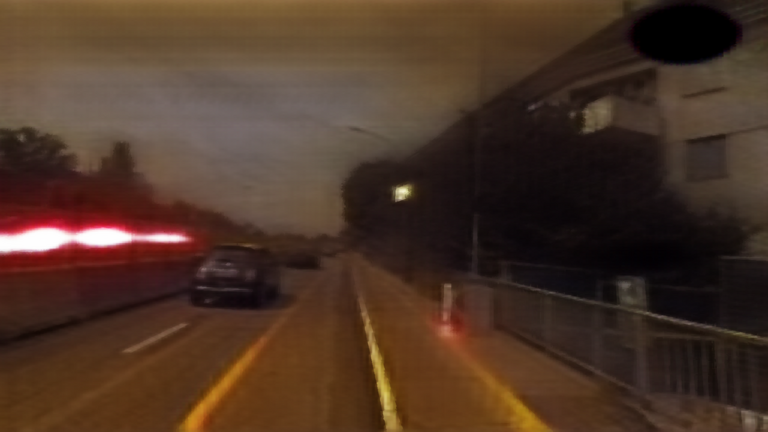}
        \caption{CycleGAN}
    \end{subfigure}
    \hfill
    \begin{subfigure}[b]{0.193\textwidth}
        \includegraphics[width=\textwidth]{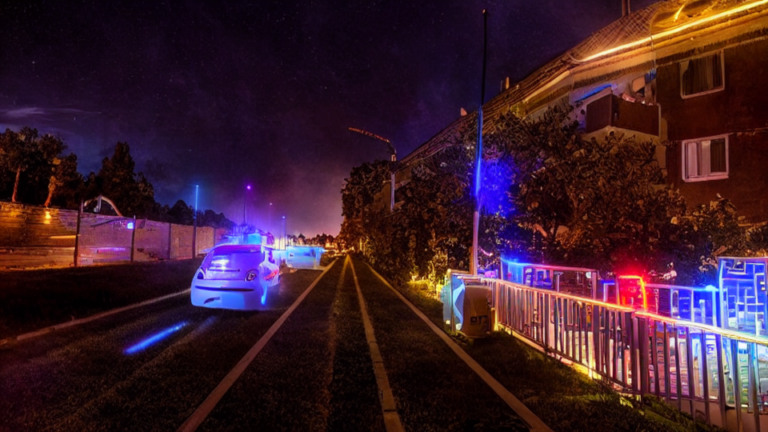}
        \caption{ControlNet}
    \end{subfigure}
    \hfill
    \begin{subfigure}[b]{0.193\textwidth}
        \includegraphics[width=\textwidth]{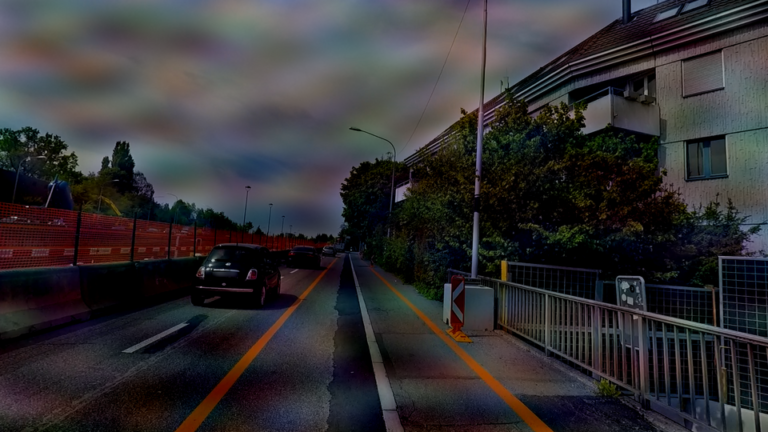}
        \caption{FDA}
    \end{subfigure}
    \hfill
    \begin{subfigure}[b]{0.193\textwidth}
        \includegraphics[width=\textwidth]{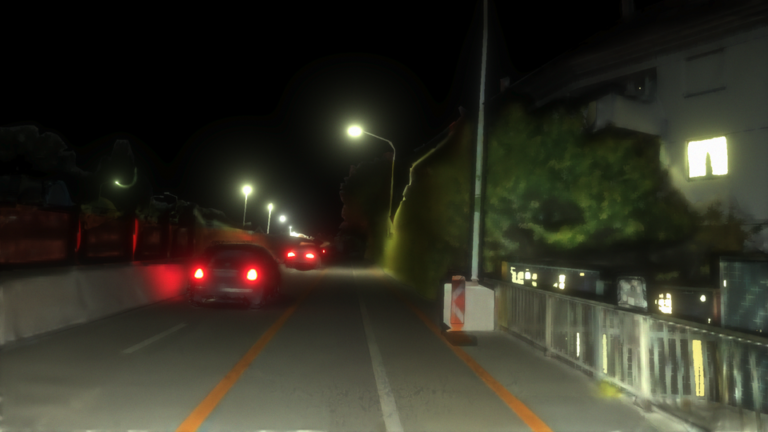}
        \caption{SOLO}
    \end{subfigure}
    
    \vspace{-0.2cm}
    \caption{\textbf{Qualitative comparison of day-to-night translation methods.} From left to right: daytime input images, and synthesized nighttime results of CycleGAN~\cite{zhu2017unpaired}, ControlNet~\cite{zhang2023adding}, FDA~\cite{yang2020fda}, and SOLO (ours).}
    \label{fig:exp:qualitative}
    \vspace{-0.2cm}
\end{figure*}
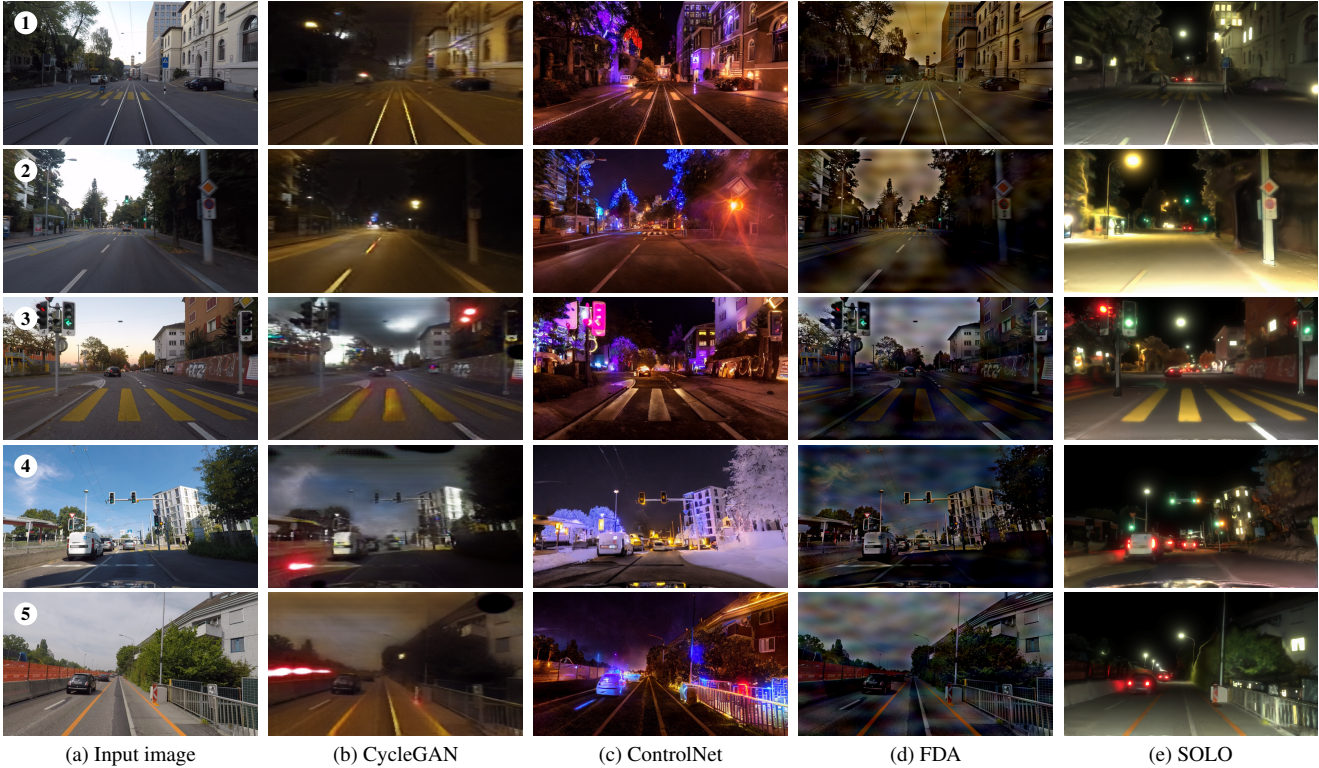

\begin{table*}[tb]
  \caption{\textbf{Comparison of day-to-night translation methods using the HRDA UDA framework for semantic segmentation on ACDC-Reference$\to$ACDC-night.} All methods are evaluated on the test split of ACDC-night.}
  \vspace{-0.3cm}
  \label{table:iou:classlevel}
  \centering
  \setlength\tabcolsep{4pt}
  \resizebox{\linewidth}{!}{%
  \begin{tabular}{l*{20}{c}}
  \toprule
  Method & \ver{road} & \ver{sidew.} & \ver{build.} & \ver{wall} & \ver{fence} & \ver{pole} & \ver{light} & \ver{sign} & \ver{veget.} & \ver{terrain} & \ver{sky} & \ver{person} & \ver{rider} & \ver{car} & \ver{truck} & \ver{bus} & \ver{train} & \ver{motorc.} & \ver{bicycle} & \textbf{mIoU}\\
  \midrule
  HRDA & 95.8 & 78.6 & 83.1 & 51.6 & 37.7 & 56.8 & 52.2 & 57.2 & 72.4 & 46.4 & 80.8 & 66.0 & 36.2 & 81.8 & 18.6 & 47.8 & 88.1 & 51.8 & 48.4 & 60.6\\
  HRDA + CycleGAN & 96.2 & 79.8 & 82.4 & 46.2 & 36.7 & 55.3 & 55.9 & 58.0 & 68.5 & 47.5 & 78.3 & 66.3 & 36.4 & 83.3 & 43.1 & 53.2 & 88.9 & 53.2 & 52.1 & 62.2\\
  HRDA + ControlNet & 95.2 & 76.9 & 80.2 & 45.0 & 27.2 & 53.2 & 55.7 & 57.9 & 54.2 & 46.0 & 66.2 & 64.0 & 39.4 & 81.3 & 56.0 & 50.3 & 88.8 & 49.9 & 47.1 & 59.7\\
  HRDA + FDA & 96.3 & 80.6 & 83.0 & 45.0 & 35.3 & 57.8 & 56.3 & 61.0 & 69.1 & 48.7 & 78.6 & 67.6 & 33.4 & 84.8 & 42.5 & 67.5 & 90.3 & 52.2 & 54.3 & 63.4\\
  HRDA + SOLO (Ours) & 95.7 & 78.8 & 82.7 & 49.4 & 31.2 & 53.7 & 51.6 & 56.8 & 71.2 & 47.6 & 78.6 & 64.3 & 36.1 & 83.2 & 65.1 & 62.3 & 89.3 & 50.9 & 48.5 & 63.0\\
  \bottomrule
  \end{tabular}}
    \vspace{-0.4cm}
\end{table*}

\subsection{Datasets}
\label{sec:exp:datasets}

In our experiments, we utilize images from ACDC~\cite{sakaridis2024acdc}, which provides panoptic annotations of the 19 Cityscapes~\cite{cordts2016cityscapes} evaluation classes for 4006 images. ACDC includes a nighttime split, further divided into training, validation and test sets. Moreover, ACDC includes daytime, clear-weather counterparts for 1003 images in the training and validation split, referred to as ACDC-Reference. We focused on ACDC for evaluating SOLO for two reasons. First, its reference split includes geographically aligned, annotated daytime counterparts of nighttime images, captured with the same camera, making it ideal for day-to-night UDA, as source and target domains only differ by the time of day. Second, due to resource limitations, annotating a large and diverse image set with light sources was infeasible, so our trained light source segmentation model may not generalize equally well to images from different sets.

\PAR{ACDC Light Sources}\,is a set contributed by this work, containing panoptic annotations of active and inactive light sources for ACDC-Reference. An initial set of 350 images was annotated manually and a light source segmentation model was then utilized for the rest 653 of the images. To this end, we fine-tuned a SegFormer model~\cite{xie2021segformer} for 40K steps using 320 annotated images as the training set.

\PAR{Nighttime Illuminants}\,is another dataset contributed by this work, derived from 60 images of a gray card illuminated by different outdoor nighttime light sources. There are five samples on average per each of the 12 light source categories of the dataset, and each sample corresponds to a chromaticity value as described in Sec.~\ref{sec:method:light_source_instantiation}.

\PAR{Evaluation dataset.}\,Images in ACDC-Reference are used both for qualitative and quantitative evaluation of SOLO.
For the latter, the state-of-the-art UDA pipeline of HRDA~\cite{hoyer2022hrda} for semantic segmentation is employed.
With the target and source domains corresponding to night time and day time respectively, a \emph{source dataset} is formed from ACDC-Reference, with 800 training and 203 validation images. Moreover, a \emph{target dataset} is formed from ACDC-night, with 400 training and 106 validation images. The 500 test images of ACDC-night with withheld labels are used as test set. The predictions of all methods on this test set are submitted to the public ACDC benchmark for evaluation.

\begin{figure*}[t]
    \centering
         
    \begin{subfigure}[b]{0.193\textwidth}
        \begin{overpic}[width=\textwidth]{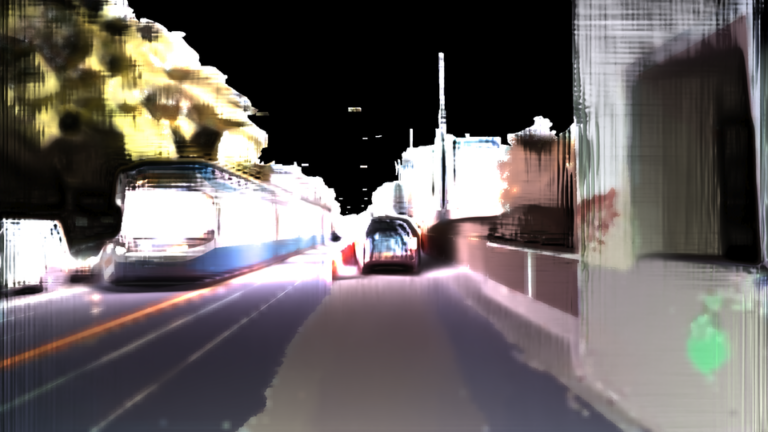}
            \put(6, 48){
                \begin{tikzpicture}[overlay]
                \node[circle, fill=white, inner sep=1pt, font=\scriptsize] (1) at (0,0) {\textbf{1}};
            \end{tikzpicture}
            }
        \end{overpic}
    \end{subfigure}
    \hfill
    \begin{subfigure}[b]{0.193\textwidth}
        \includegraphics[width=\textwidth]{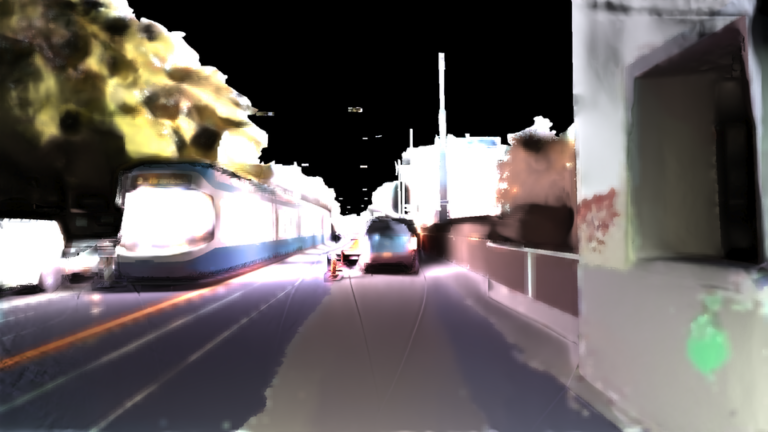}
    \end{subfigure}
    \hfill
    \begin{subfigure}[b]{0.193\textwidth}
        \includegraphics[width=\textwidth]{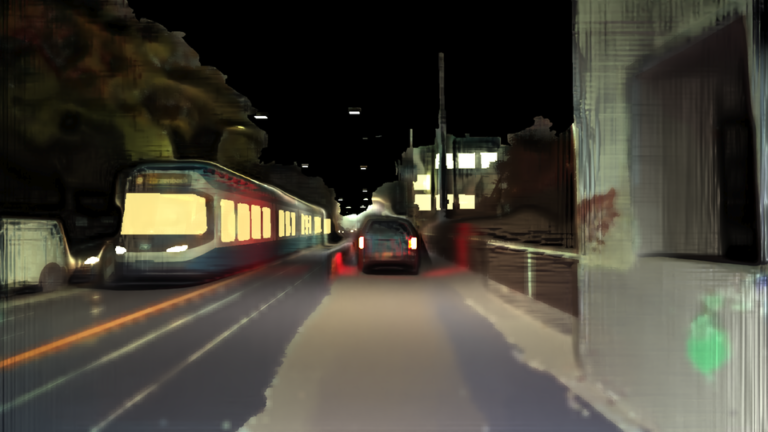}
    \end{subfigure}
    \hfill
    \begin{subfigure}[b]{0.193\textwidth}
        \includegraphics[width=\textwidth]{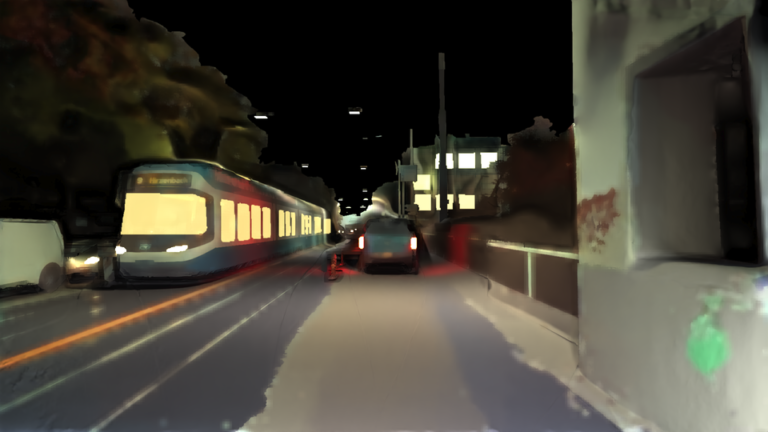}
    \end{subfigure}
    \hfill
    \begin{subfigure}[b]{0.193\textwidth}
        \includegraphics[width=\textwidth]{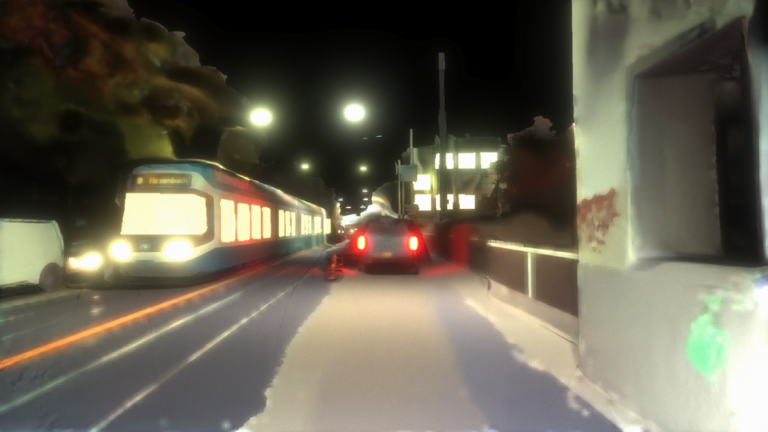}
    \end{subfigure}
       
    \vspace{0.2em}
        
    \begin{subfigure}[b]{0.193\textwidth}
        \begin{overpic}[width=\textwidth]{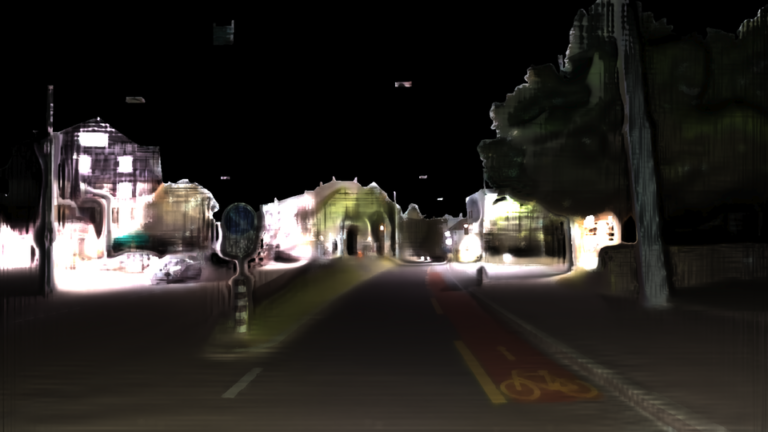}
            \put(6, 48){
                \begin{tikzpicture}[overlay]
                \node[circle, fill=white, inner sep=1pt, font=\scriptsize] (1) at (0,0) {\textbf{2}};
            \end{tikzpicture}
            }
        \end{overpic}
        \caption{ablation 0}
        \label{fig:exp:ablation:0}
    \end{subfigure}
    \hfill
    \begin{subfigure}[b]{0.193\textwidth}
        \includegraphics[width=\textwidth]{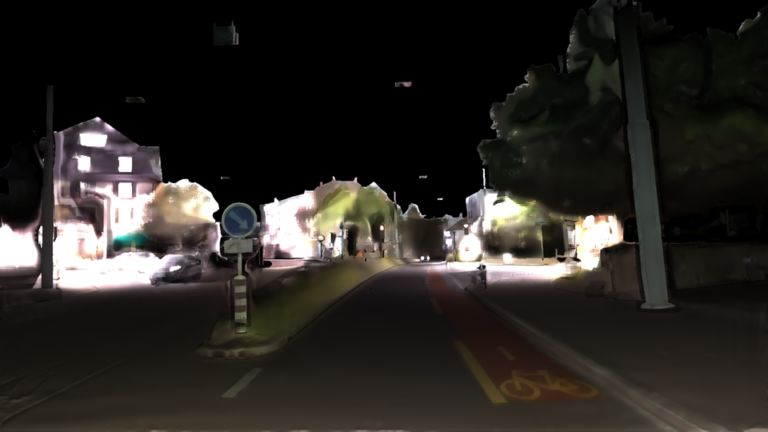}
        \caption{ablation 1}
        \label{fig:exp:ablation:1}
    \end{subfigure}
    \hfill
    \begin{subfigure}[b]{0.193\textwidth}
        \includegraphics[width=\textwidth]{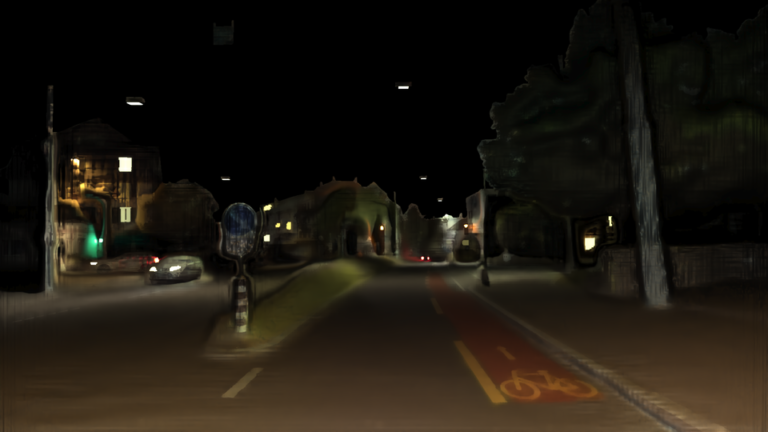}
        \caption{ablation 2}
        \label{fig:exp:ablation:2}
    \end{subfigure}
    \hfill
    \begin{subfigure}[b]{0.193\textwidth}
        \includegraphics[width=\textwidth]{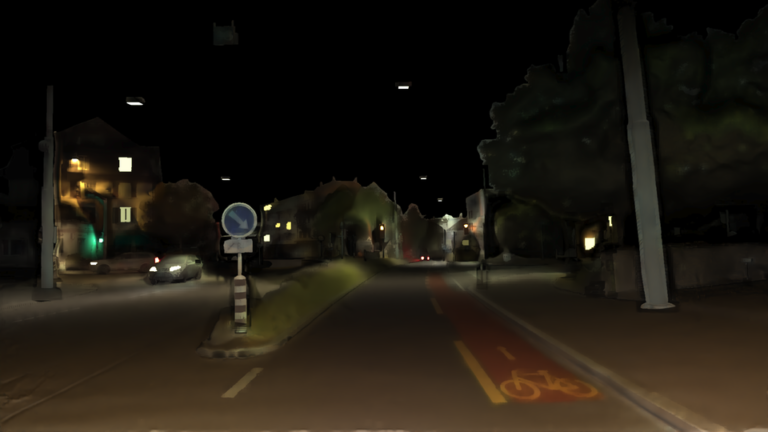}
        \caption{ablation 3}
        \label{fig:exp:ablation:3}
    \end{subfigure}
    \hfill
    \begin{subfigure}[b]{0.193\textwidth}
        \includegraphics[width=\textwidth]{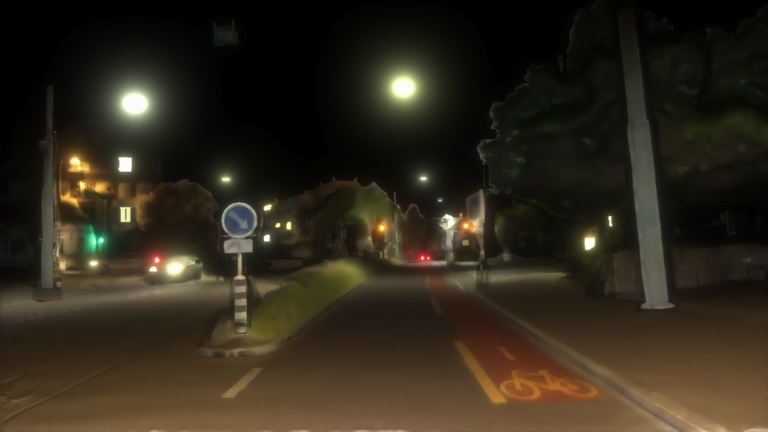}
        \caption{SOLO}
        \label{fig:exp:ablation:solo}
    \end{subfigure}
    
    \vspace{-0.2cm}
    \caption{\textbf{Ablation study of SOLO.} From column (a) - (d) the results of the ablated versions of SOLO i.e.\ 0, 1, 2, and 3 are presented. SOLO generated nighttime images are displayed in column (e). Notably, every sample (row) is labeled with a number.}
    \label{fig:exp:ablation}
    \vspace{-0.6cm}
\end{figure*}

\subsection{Comparisons to The State of The Art}
\label{sec:exp:comparisons}

SOLO is compared against other state-of-the-art stylization methods, including Fourier Domain Adaptation (FDA)~\cite{yang2020fda}, ControlNet~\cite{zhang2023adding} and CycleGAN~\cite{zhu2017unpaired}. 
For FDA, a bandwidth of 0.01 is set. For ControlNet, the prompt 'transform this image to nighttime' is used, showing little difference when paraphrased. For the \emph{qualitative comparison}, the generated stylized images alongside their daytime inputs are displayed in Fig.~\ref{fig:exp:qualitative}.
In column (b), CycleGAN generations exhibit several issues including: the incomplete removal of the daytime ambient illumination (images 3 and 4), the unrealistic, light blue, appearance of the nighttime sky,
the spatially inconsistent red light glows (images 3, 4 and 5), the inactive vehicle and traffic lights, and the unrealistic illumination of regions that surround activated light sources as though those sources were inactive. By contrast, ControlNet (column (c)) achieves a more realistic rendering of the nighttime sky. However, traffic (images 3 and 4) and street (images 2 and 5) lights remain inactive, similar to CycleGAN. Additionally, a strong violet tint is present across the stylized images, and the surface textures are not inherited from the daytime inputs. FDA (column (d)) faces similar challenges to CycleGAN, including the incomplete elimination of the ambient light, the unrealistic color of the nighttime sky, a repetitive gray pattern in sky regions, and the failure to account for light sources, resulting in an unrealistic nighttime result. On the other hand, SOLO (column (e)) tackles most of the aforementioned issues. The ambient illumination from the daytime image is eliminated, scene illumination fully depends on the activated light sources, and the surface textures from the daytime image are closely resembled. Notably, the instantiation of the light sources is explicitly handled. The colors of the lights are sampled from the nighttime illuminants dataset, conditioned to the inherited daytime semantics. The noise addition and fog glare effects realistically simulate various typical nighttime artifacts.

SOLO is quantitatively evaluated using the HRDA~\cite{hoyer2022hrda} UDA framework. In particular, the `MiT-B5' SegFormer semantic segmentation model \cite{xie2021segformer}, pre-trained on daytime images, is adapted to the target nighttime domain.
Moreover, three random seeds are used for each setting during training. The mean intersection-over-union (mIoU) is used to select the best model for evaluation on the test set of ACDC-night. The test mIoU and class-level IoUs are reported. In Table~\ref{table:iou:classlevel}, SOLO outperforms both state-of-the-art input-level adaptation methods including CycleGAN and ControlNet and the original HRDA and achieves an mIoU score of $63.0\%$. However, FDA slightly outperforms SOLO, despite the visually inferior qualitative results of the former in Fig.~\ref{fig:exp:qualitative}.
We hypothesize that this quantitative UDA-based comparison is not suitable to fully demonstrate the superior realism of the nighttime images rendered with SOLO. This is because images rendered with SOLO are generally darker than those output by FDA.
Consequently, as dark regions in SOLO images appearing as coherent segments may actually include segments from different classes, the inherited daytime annotations may not correspond to discernible segments and thus confuse the model.

\subsection{Ablation Study}
An ablation study is conducted to evaluate each component of our method. Four ablated versions are presented, evaluated both qualitatively and quantitatively. For the former, stylized images of SOLO and its ablations are shown in Fig.~\ref{fig:exp:ablation}. To verify the qualitative observations, the HRDA framework for semantic segmentation is employed in Table~\ref{table:exp:ablation}. The ablated versions are formed by `switching off' component sets of SOLO. Essential components, such as backprojection, are not ablated, but the rest are divided into geometric, light instantiation and image post-processing sets. Disabling the geometric set removes the instance-reference cross-bilateral filter, the normal-guided depth optimization and the mesh post-processing. Similarly, ablating the light instantiation set assumes all light sources are active, sets all light source chromaticities to white, and uses a uniform strength value. Finally, by switching off the image post-processing set, the noise addition and the fog glare effect are disabled.

\begin{table}[tb]
  \caption{\textbf{Ablation study of SOLO using HRDA framework for the semantic
segmentation task.}
  ``geometric'': set of geometric components, ``lights inst.'': set of lights instantiation components, ``image post-proc.'': set of image post-processing components.
  }
  \vspace{-0.2cm} 
  \centering
  \setlength\tabcolsep{4pt}
  \resizebox{\linewidth}{!}{%
  \begin{tabular}{lcccc}
  \toprule
  id & geometric  & lights inst. & image post-proc. & mIoU \\
  \midrule
  0 & \no         & \no        & \no             & 52.7 $\pm$ 1.6\\
  1 & \yes        & \no        & \no             & 53.5 $\pm$ 1.0 \\
  2 & \no         & \yes       & \no             & 52.4 $\pm$ 0.3 \\
  3 & \yes        & \yes       & \no             & 53.6 $\pm$ 0.9 \\
  \midrule
  4 (SOLO) & \yes & \yes       & \yes            & 55.1 $\pm$ 0.4 \\
  \bottomrule
  \end{tabular}}
  \label{table:exp:ablation}
  \vspace{-0.5cm}
\end{table}

In Table~\ref{table:exp:ablation}, SOLO outperforms all the ablated versions significantly. Significant difference in mIoU is also observed when either the image post-processing (SOLO $\rightarrow$ ablation 3) or the geometric (ablation 3 $\rightarrow$ ablation 2) component sets are ablated. These differences are also evident in the qualitative results of Fig.~\ref{fig:exp:ablation}. Specifically, artifacts in the geometry are greatly reduced when the geometric component set is included (ablation 0 $\rightarrow$ ablation 1) and the image post-processing components result in more realistic renderings, introducing artifacts typical at night time (ablation 3 $\rightarrow$ SOLO). However, the quantitative results for the lights instantiation ablation are inconclusive. We attribute this finding to the increased brightness of renderings without our light instantiation components, as all light sources are activated for these, as opposed to partial activation with our method (cf.\ Fig.~\ref{fig:exp:ablation:1} vs.\ \ref{fig:exp:ablation:3}). This increase leads to more discernible objects, which counteracts the reduced realism when it comes to semantic segmentation performance.

\section{Conclusion}

We present SOLO, the first monocular, physically-based method for simulating photorealistic nighttime versions of daytime scenes. Our method features several novel contributions, such as a probabilistic light source instantiation module which selectively activates light sources in the scene to achieve more realistic and contextually accurate results. Moreover, we employ a pipeline guided by semantics to fuse geometric representations into a single 3D mesh for usage in forward rendering. Our image post-processing pipeline effectively mimics typical camera artifacts for night time. Our results suggest that SOLO significantly outperforms current state-of-the-art data-driven time-of-day-transfer approaches in the context of day-to-night UDA, highlighting the importance of semantic and physically-based priors in synthesizing photorealistic nighttime images. Finally, we believe that our ACDC Light Sources and Nighttime Illuminants datasets will be valuable resources for the community in working on night time.

\section*{Acknowledgment}
This work was supported by an ETH Career Seed Award.

{\small
\bibliographystyle{ieee_fullname}
\bibliography{refs}
}

\clearpage

\appendix

\begin{table*}[tb]
\caption{\textbf{ACDC Light Sources dataset labels.} The annotation and prediction labels along with a short description are organized here.}
\centering
\resizebox{\linewidth}{!}{%
\begin{tabular}{| l | c | c |}
\hline
\textbf{Description} & \textbf{Annotation labels} & \textbf{Prediction labels} \\
\hline
Windows of buildings & window\_building & window\_building \\
\hline
Windows of parked vehicles & window\_parked & window\_parked  \\
\hline
Windows of public transport vehicles (e.g.\ bus, tram, etc.) & window\_transport & window\_transport  \\
\hline
Traffic lights & traffic\_light & traffic\_light  \\
\hline
Street lights of high correlated color temperature & street\_light\_HT & \multirow{2}{4em}{street\_light} \\
\cline{1-2}
Street lights of low correlated color temperature & street\_light\_LT & \\
\hline
Front lights of parked vehicles & parked\_front & \multirow{2}{4em}{front\_light} \\
\cline{1-2}
Front lights of moving vehicles & moving\_front & \\
\hline
Rear lights of parked vehicles & parked\_rear & \multirow{2}{4em}{rear\_light} \\
\cline{1-2}
Rear lights of moving vehicles & moving\_rear & \\
\hline
Emitting advertisement panels & advertisement & advertisement  \\
\hline
Clocks (e.g. such as those near bus stops) & clock & \multirow{2}{4em}{inferred} \\
\cline{1-2}
Lights whose light color can be inferred from the daytime image & inferred & \\
\hline
Group of windows located in the same building floor & windows\_group & \no \\
\hline
\end{tabular}
}
\label{table:labels}
\end{table*}

\section{ACDC Light Sources Dataset Labels}
\label{sec:sup:light-labels}
As described in Sec.~\ref{sec:method:light_source_instantiation}, to create the ACDC Light Sources dataset we first manually annotated a small subset of it. A semantic segmentation network was then trained on the manually annotated data to predict the semantic masks for the rest of the images in the dataset. The labels used initially were constrained by the ones the network could accurately predict. For example, ``parked\_front" and ``moving\_front" that correspond to the front lights of a vehicle when it is parked or moving respectively, are merged into one label. A complete list of the labels used both for annotation and prediction is displayed in Table~\ref{table:labels}.

\section{
 Qualitative Results}
\label{sec:sup:qualitative}
\begin{figure*}[h]
    \centering
    \begin{subfigure}[b]{0.193\textwidth}
        \begin{overpic}[width=\textwidth]{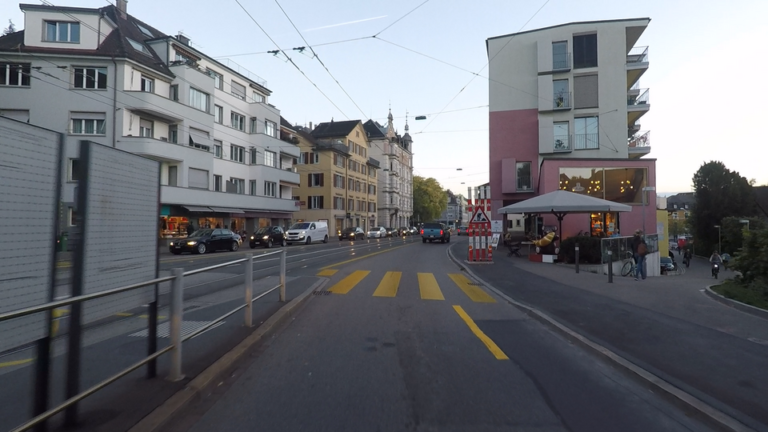}
            \put(6, 48){
                \begin{tikzpicture}[overlay]
                \node[circle, fill=white, inner sep=1pt, font=\scriptsize] (1) at (0,0) {\textbf{1}};
            \end{tikzpicture}
            }
        \end{overpic}
    \end{subfigure}
    \hfill
    \begin{subfigure}[b]{0.193\textwidth}
        \includegraphics[width=\textwidth]{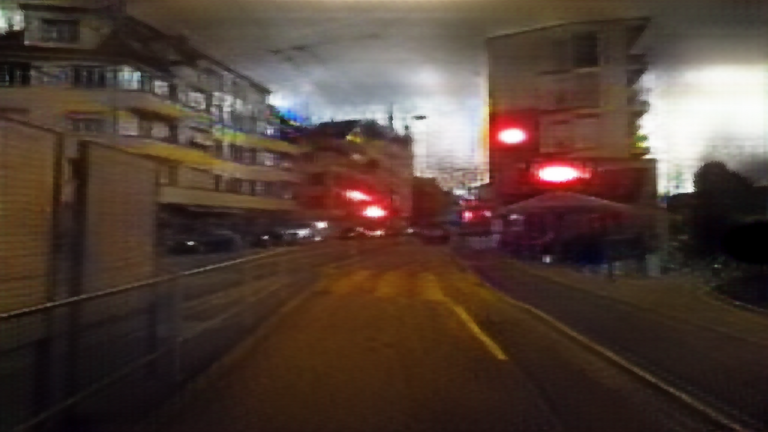}
    \end{subfigure}
    \hfill
    \begin{subfigure}[b]{0.193\textwidth}
        \includegraphics[width=\textwidth]{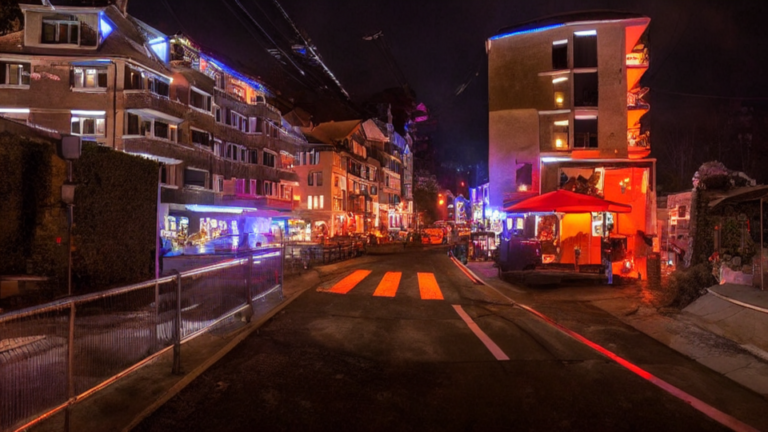}
    \end{subfigure}
    \hfill
    \begin{subfigure}[b]{0.193\textwidth}
        \includegraphics[width=\textwidth]{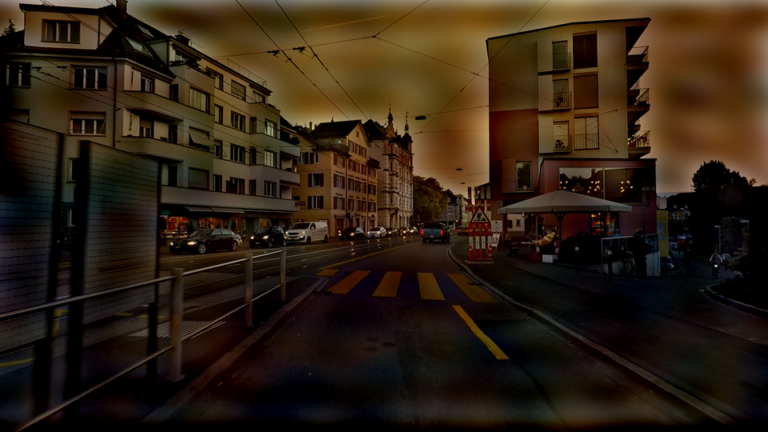}
    \end{subfigure}
    \hfill
    \begin{subfigure}[b]{0.193\textwidth}
        \includegraphics[width=\textwidth]{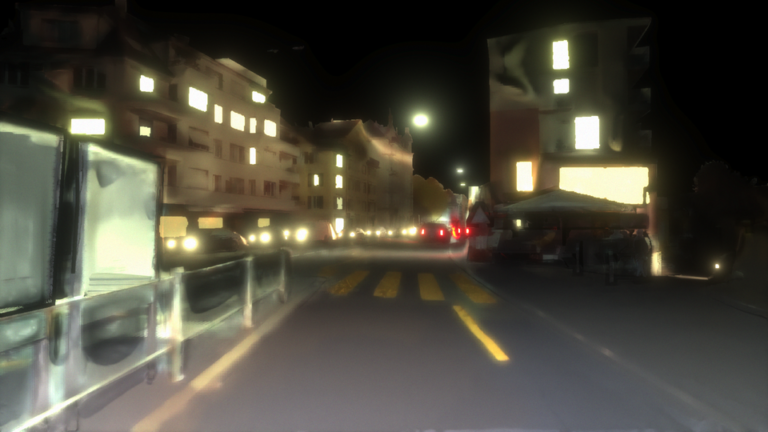}
    \end{subfigure}
    
    \vspace{0.2em}
        
    \begin{subfigure}[b]{0.193\textwidth}
        \begin{overpic}[width=\textwidth]{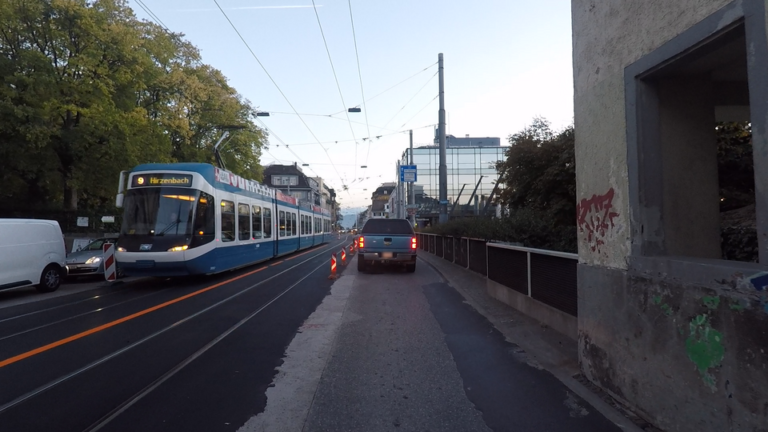}
            \put(6, 48){
                \begin{tikzpicture}[overlay]
                \node[circle, fill=white, inner sep=1pt, font=\scriptsize] (1) at (0,0) {\textbf{2}};
            \end{tikzpicture}
            }
        \end{overpic}
    \end{subfigure}
    \hfill
    \begin{subfigure}[b]{0.193\textwidth}
        \includegraphics[width=\textwidth]{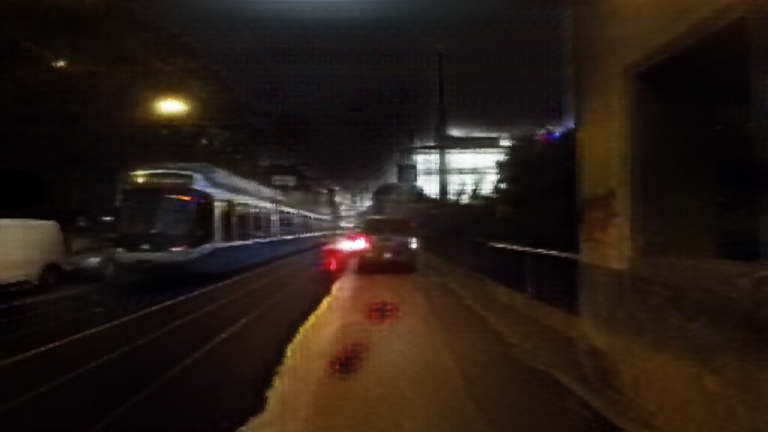}
    \end{subfigure}
    \hfill
    \begin{subfigure}[b]{0.193\textwidth}
        \includegraphics[width=\textwidth]{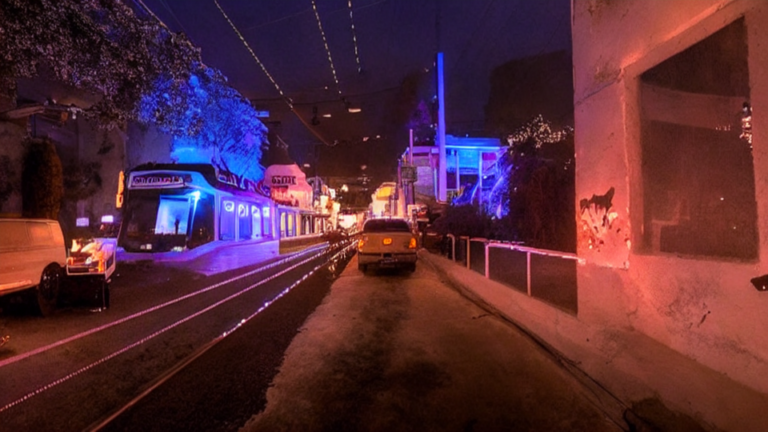}
    \end{subfigure}
    \hfill
    \begin{subfigure}[b]{0.193\textwidth}
        \includegraphics[width=\textwidth]{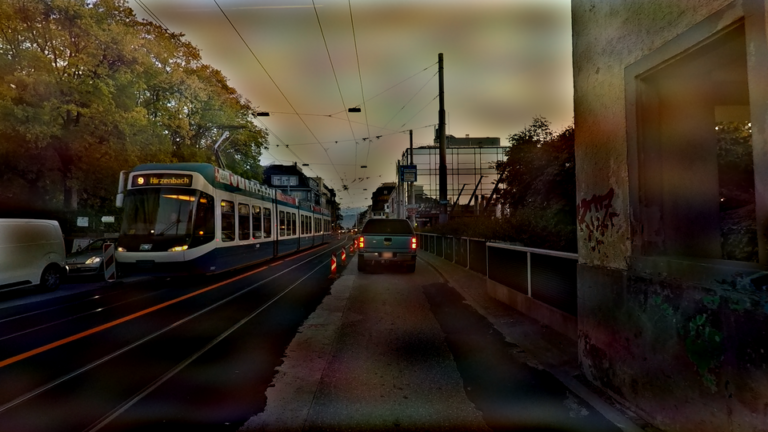}
    \end{subfigure}
    \hfill
    \begin{subfigure}[b]{0.193\textwidth}
        \includegraphics[width=\textwidth]{figures/samples/solov3/GOPR0351_frame_000515.png}
    \end{subfigure}
    
    \vspace{0.2em}
        
    \begin{subfigure}[b]{0.193\textwidth}
        \begin{overpic}[width=\textwidth]{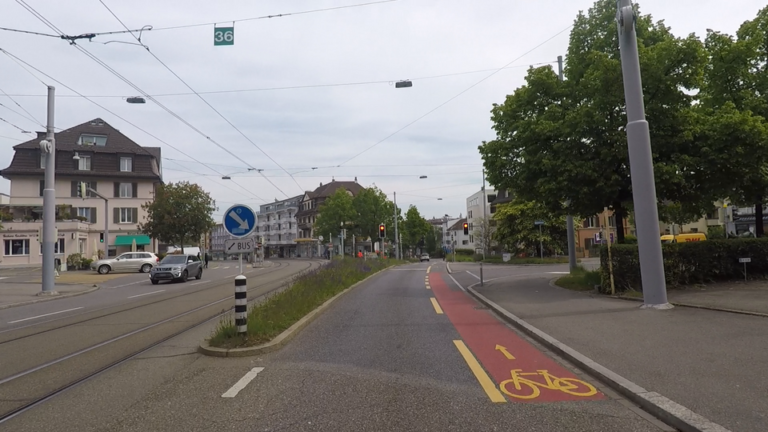}
            \put(6, 48){
                \begin{tikzpicture}[overlay]
                \node[circle, fill=white, inner sep=1pt, font=\scriptsize] (1) at (0,0) {\textbf{3}};
            \end{tikzpicture}
            }
        \end{overpic}
    \end{subfigure}
    \hfill
    \begin{subfigure}[b]{0.193\textwidth}
        \includegraphics[width=\textwidth]{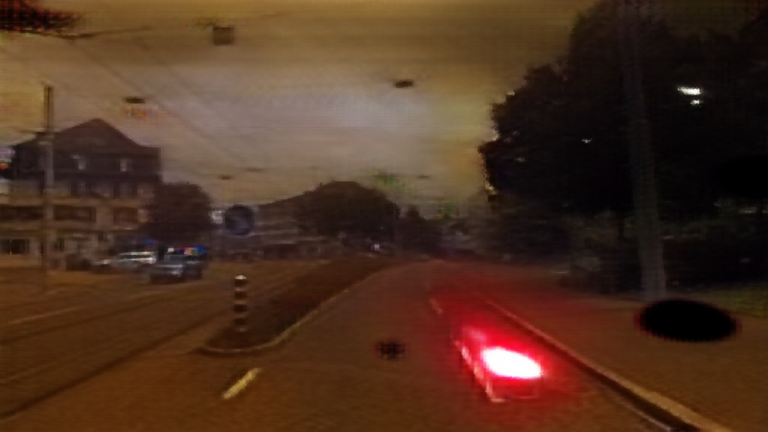}
    \end{subfigure}
    \hfill
    \begin{subfigure}[b]{0.193\textwidth}
        \includegraphics[width=\textwidth]{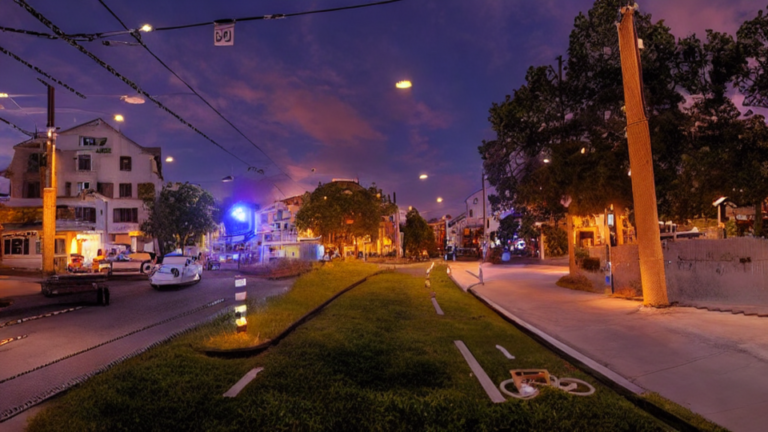}
    \end{subfigure}
    \hfill
    \begin{subfigure}[b]{0.193\textwidth}
        \includegraphics[width=\textwidth]{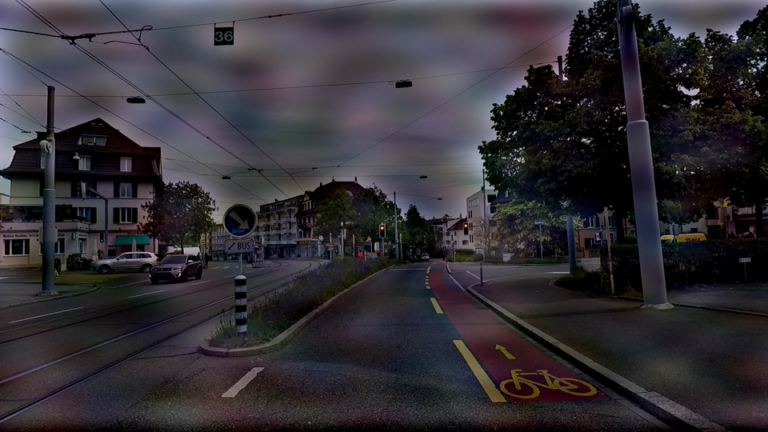}
    \end{subfigure}
    \hfill
    \begin{subfigure}[b]{0.193\textwidth}
        \includegraphics[width=\textwidth]{figures/samples/solov3/GOPR0476_frame_000481.png}
    \end{subfigure}
    
    \vspace{0.2em}
        
    \begin{subfigure}[b]{0.193\textwidth}
        \begin{overpic}[width=\textwidth]{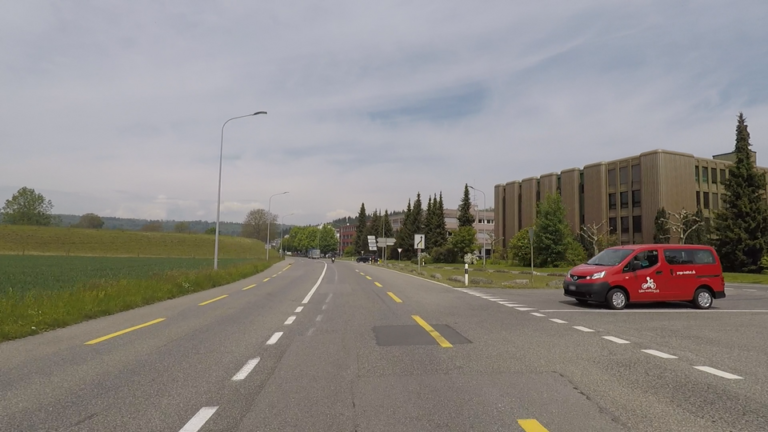}
            \put(6, 48){
                \begin{tikzpicture}[overlay]
                \node[circle, fill=white, inner sep=1pt, font=\scriptsize] (1) at (0,0) {\textbf{4}};
            \end{tikzpicture}
            }
        \end{overpic}
        \caption{Input image}
    \end{subfigure}
    \hfill
    \begin{subfigure}[b]{0.193\textwidth}
        \includegraphics[width=\textwidth]{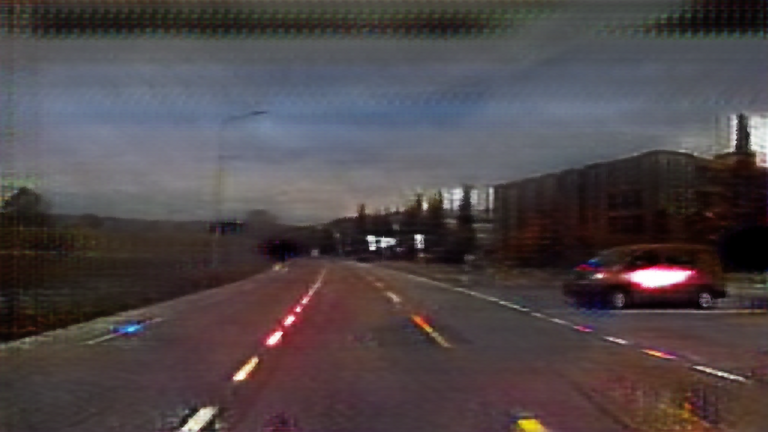}
        \caption{CycleGAN}
    \end{subfigure}
    \hfill
    \begin{subfigure}[b]{0.193\textwidth}
        \includegraphics[width=\textwidth]{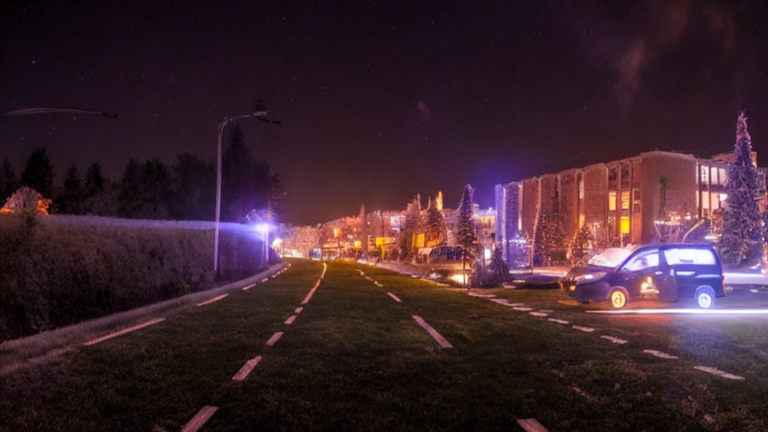}
        \caption{ControlNet}
    \end{subfigure}
    \hfill
    \begin{subfigure}[b]{0.193\textwidth}
        \includegraphics[width=\textwidth]{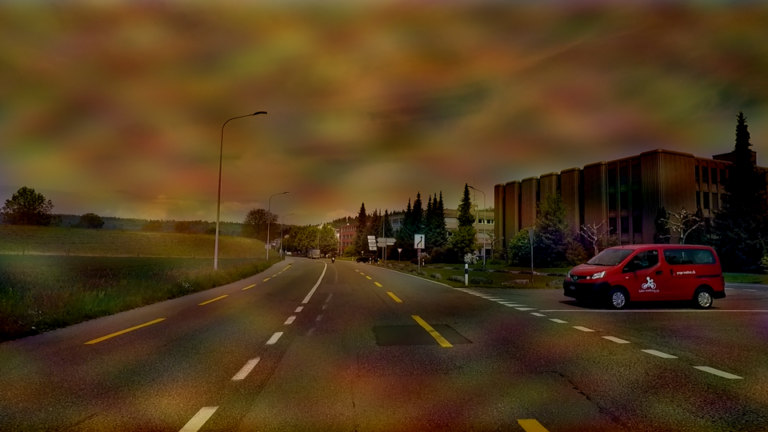}
        \caption{FDA}
    \end{subfigure}
    \hfill
    \begin{subfigure}[b]{0.193\textwidth}
        \includegraphics[width=\textwidth]{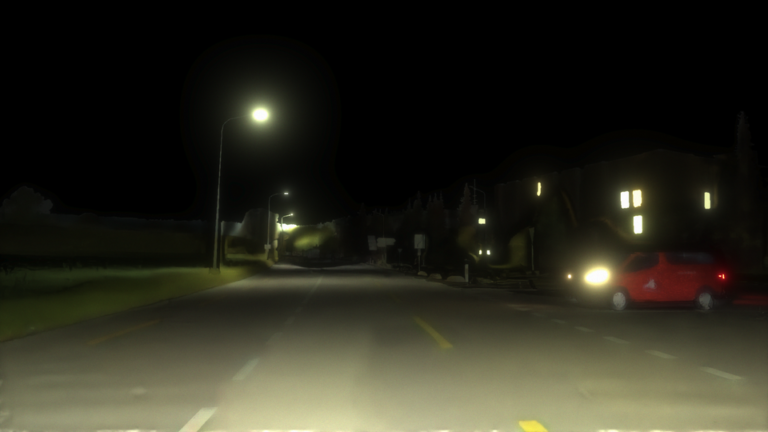}
        \caption{SOLO}
    \end{subfigure}
    
    \vspace{-0.2cm}
    \caption{\textbf{Additional qualitative comparisons of day-to-night translation methods.} From left to right: daytime input images, and synthesized nighttime results of CycleGAN~\cite{zhu2017unpaired}, ControlNet~\cite{zhang2023adding}, FDA~\cite{yang2020fda}, and SOLO (ours).}
    \label{fig:sup:qualitative}
    \vspace{-0.2cm}
\end{figure*}

Some additional visual results for the qualitative comparison of day-to-night translation methods presented in Sec.~\ref{sec:exp:comparisons} are illustrated in Fig.~\ref{fig:sup:qualitative}.

\section{Nighttime Illuminants Dataset Samples}
\label{sec:sup:illuminants}
As described in Sec.~\ref{sec:method:light_source_instantiation}, for the collection of the Nighttime Illuminants dataset, raw images of a gray card positioned under the illuminant of interest were captured. A gray card is a diffuse surface of known and fixed spectral reflectance across the visible spectrum. In our setting, a 20x25cm Kodak Gray Card reflecting 18\% of the light across the visible spectrum was used. These images are then processed by an image processing pipeline to acquire the corresponding chromaticity coordinates. First, the area of the collected images to be processed was manually annotated. Second, the metadata of the raw images stored in the Adobe DNG format \footnote{https://helpx.adobe.com/camera-raw/digital-negative.html} were parsed, including active sensor's area, Bayer filter pattern, black and white levels and the camera color matrices. The pipeline continues by filtering the inactive sensor's pixels and applying normalization and black level subtraction. After that, the image is white-balanced using the CIE's \footnote{https://cie.co.at/publications/colorimetric-illuminants} standard illuminant E. The demosaicing process follows, along with a transformation from the color space of the camera to the XYZ color space. The chromaticity coordinates $(x,y)$ are finally computed from the average $(X, Y, Z)$ coordinates of the annotated image region.

\section{Probabilistic Instantiation Uniform Bounds}
\label{sec:sup:instantiation}
In Sec.~\ref{sec:method:light_source_instantiation}, the light source instantiation module has been described. In Table~\ref{table:uniform-bounds} we additionally provide the empirically selected parameters (i.e.\ bounds) for the discrete uniform distribution from which the Bernoulli parameters that control the activation of the light source are sampled from. The values of those parameters are also conditioned to the light source group. This group is selected based on the panoptic information of the reference images of the ACDC dataset.

\begin{table*}[tb]
\caption{\textbf{Empirically set intervals} for uniform distributions from which the Bernoulli parameters in light source activation are sampled.}
\label{table:uniform-bounds}
\centering
\begin{tabular}{| c | c | c|}
\hline
\textbf{light source group} & \textbf{light source} & \textbf{uniform bounds}\\
        \hline
            \multirow{9}{*}{N/A} 
                 & inferred          & (1, 1) \\ \cline{2-3}
                & traffic\_light\_G & (1, 1) \\ \cline{2-3}
                & traffic\_light\_R & (1, 1) \\ \cline{2-3}
                & traffic\_light\_O & (1, 1) \\ \cline{2-3}
                & street\_light\_HT & (1, 1) \\ \cline{2-3}
                & street\_light\_LT & (1, 1) \\ \cline{2-3}
                & advertisement     & (.6, .8) \\ \cline{2-3}
                & clock             & (.8, 1) \\ \cline{2-3}
                & window\_building  & (.3, .6) \\ \cline{2-3}
            \hline
            building floor    & window\_building & (.3, .6) \\ \cline{2-3}
            \hline
            \multirow{5}{*}{car} & window\_parked & (.1, .4) \\ \cline{2-3}
                     & moving\_front    & (.95, 1) \\ \cline{2-3}
                     & moving\_rear     & (.95, 1) \\ \cline{2-3}
                     & parked\_front    & (.1, .3) \\ \cline{2-3}
                     & parked\_rear     & (.1, .3) \\ \cline{2-3}
            \hline
            \multirow{7}{*}{bus} & window\_parked & (.1, .4) \\ \cline{2-3}
                     & moving\_front & (.95, 1) \\ \cline{2-3}
                     & moving\_rear & (.95, 1) \\ \cline{2-3}
                     & parked\_front & (.1, .3) \\ \cline{2-3}
                     & parked\_rear & (.1, .3) \\ \cline{2-3}
                     & window\_transport & (.9, 1) \\ \cline{2-3}
                     & inferred & (1, 1) \\ \cline{2-3}
            \hline
            \multirow{6}{*}{tram} & moving\_front & (.95, 1) \\ \cline{2-3}
                     & moving\_rear & (.95, 1) \\ \cline{2-3}
                     & parked\_front & (.1, .3) \\ \cline{2-3}
                     & parked\_rear & (.1, .3) \\ \cline{2-3}
                     & window\_transport & (.9, 1) \\ \cline{2-3}
                     & inferred & (1, 1) \\ \cline{2-3}
            \hline
            \multirow{5}{*}{truck} & window\_parked & (.1, .4) \\ \cline{2-3}
                     & moving\_front & (.95, 1) \\ \cline{2-3}
                     & moving\_rear & (.95, 1) \\ \cline{2-3}
                     & parked\_front & (.1, .3) \\ \cline{2-3}
                     & parked\_rear & (.1, .3) \\ \cline{2-3}
            \hline
            \multirow{4}{*}{motorcycle} & moving\_front & (.95, 1) \\ \cline{2-3}
                     & moving\_rear & (.95, 1) \\ \cline{2-3}
                     & parked\_front & (.1, .3) \\ \cline{2-3}
                     & parked\_rear & (.1, .3) \\ \cline{2-3}
            \hline
            \multirow{4}{*}{bicycle} & moving\_front & (.95, 1) \\ \cline{2-3}
                     & moving\_rear & (.95, 1) \\ \cline{2-3}
                     & parked\_front & (.1, .2) \\ \cline{2-3}
                     & parked\_rear & (.1, .2) \\
        
\hline
\end{tabular}
\label{table:categories}
\end{table*}

\begin{figure*}[h]
    \centering
    \begin{subfigure}[b]{0.193\textwidth}
        \includegraphics[width=\textwidth]{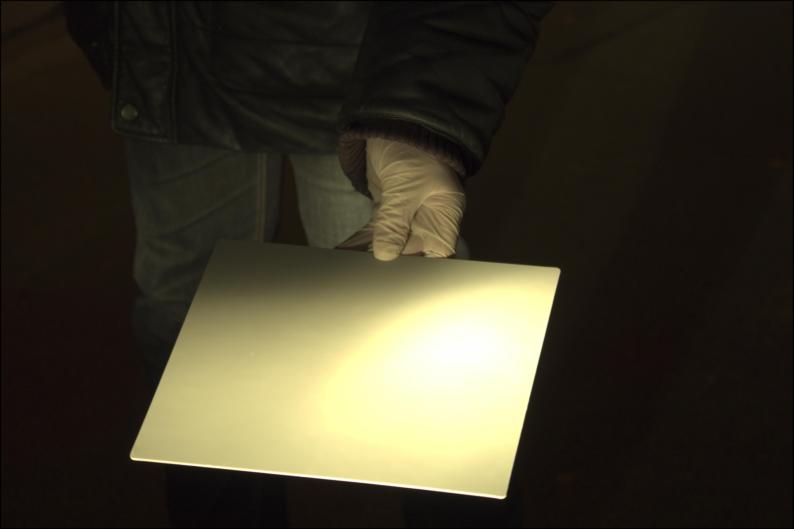}
    \end{subfigure}
    \hfill
    \begin{subfigure}[b]{0.193\textwidth}
        \includegraphics[width=\textwidth]{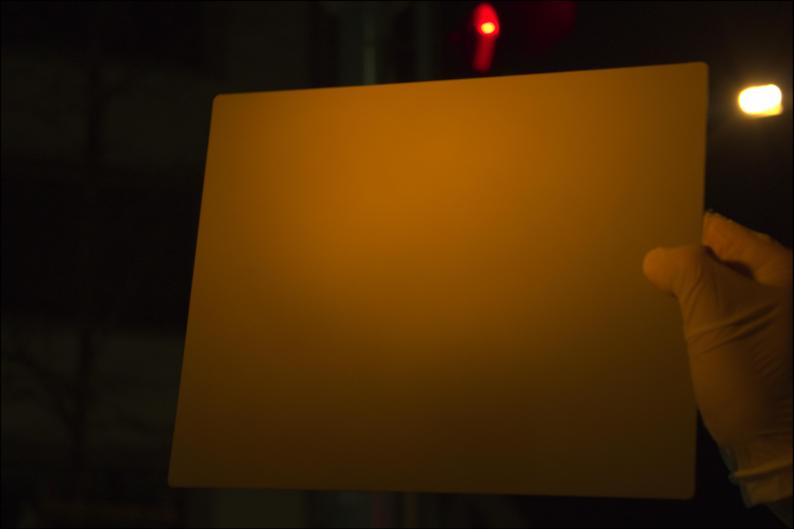}
    \end{subfigure}
    \hfill
    \begin{subfigure}[b]{0.193\textwidth}
        \includegraphics[width=\textwidth]{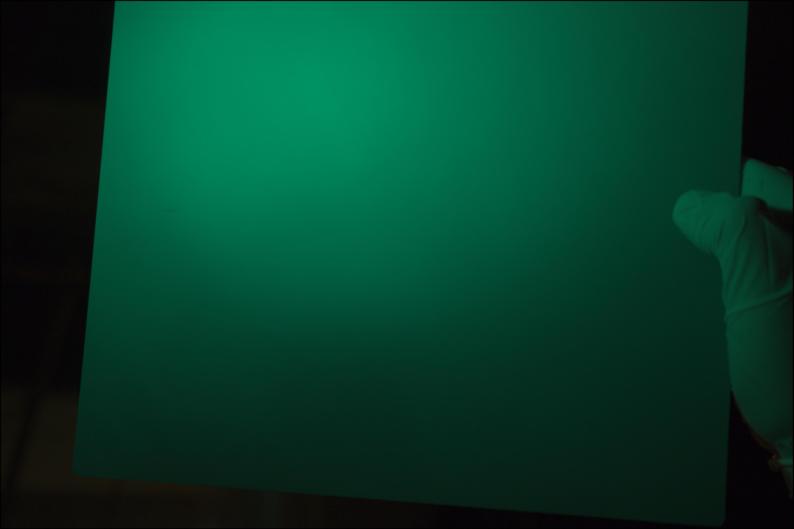}
    \end{subfigure}
    \hfill
    \begin{subfigure}[b]{0.193\textwidth}
        \includegraphics[width=\textwidth]{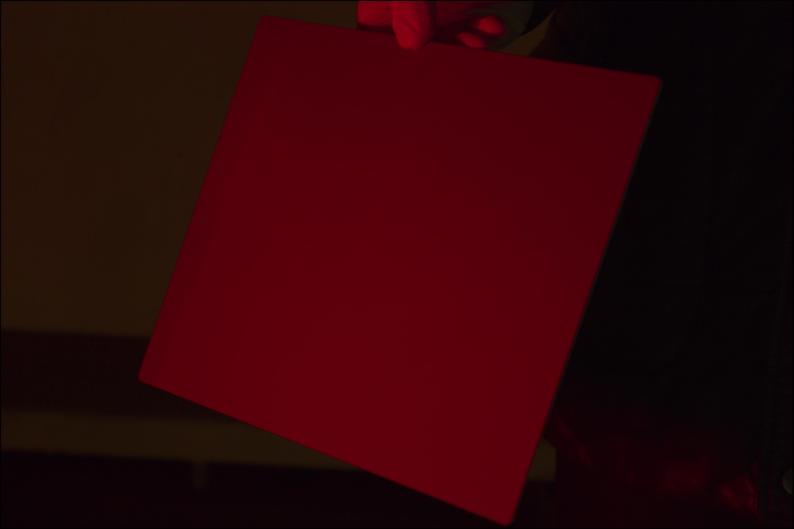}
    \end{subfigure}
    \hfill
    \begin{subfigure}[b]{0.193\textwidth}
        \includegraphics[width=\textwidth]{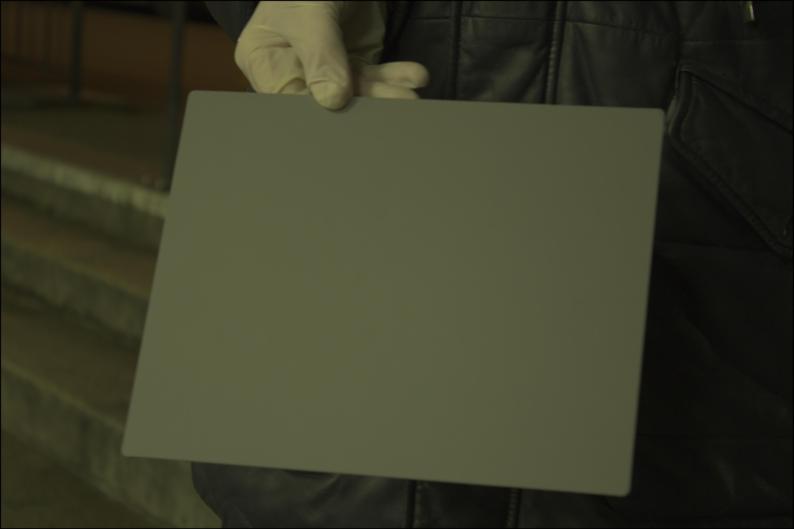}
    \end{subfigure}
    
    \vspace{0.2em}
    
    \begin{subfigure}[b]{0.193\textwidth}
        \includegraphics[width=\textwidth]{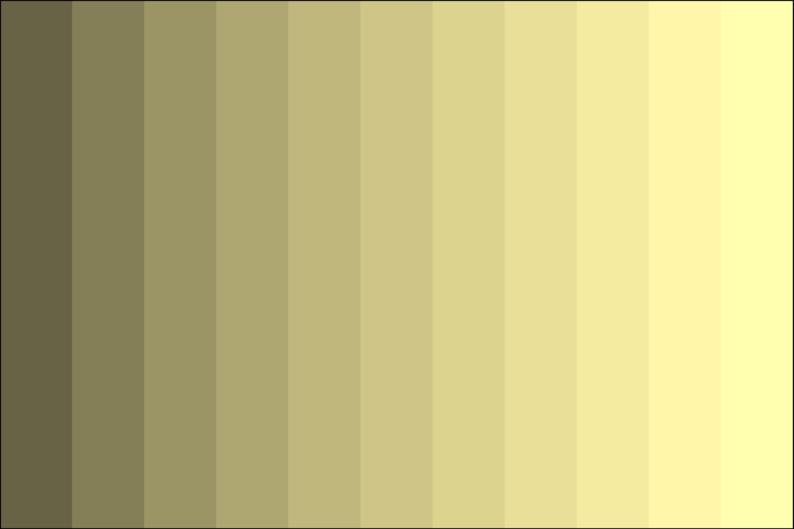}
    \end{subfigure}
    \hfill
    \begin{subfigure}[b]{0.193\textwidth}
        \includegraphics[width=\textwidth]{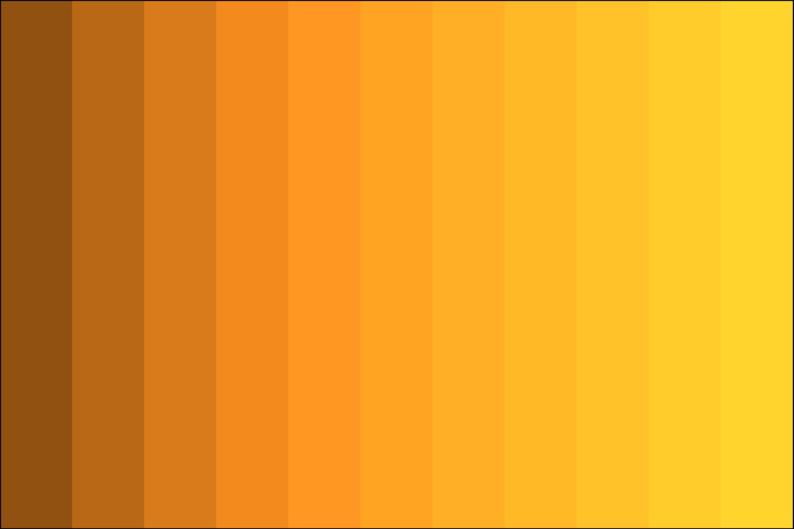}
    \end{subfigure}
    \hfill
    \begin{subfigure}[b]{0.193\textwidth}
        \includegraphics[width=\textwidth]{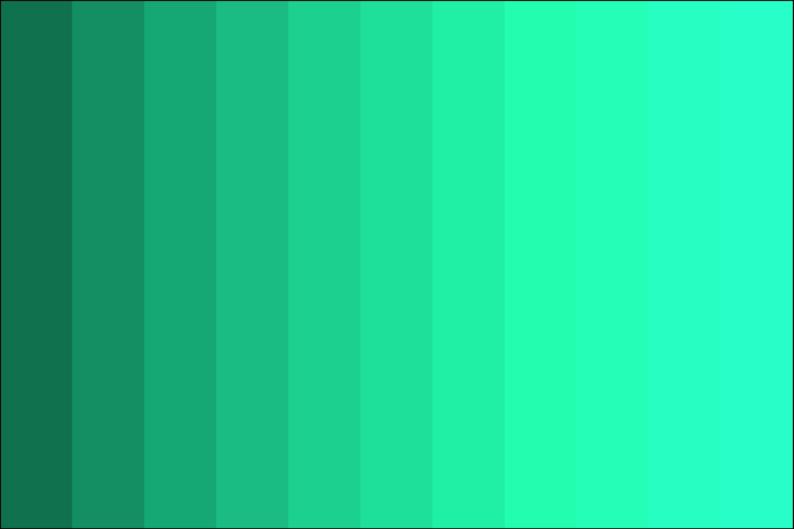}
    \end{subfigure}
    \hfill
    \begin{subfigure}[b]{0.193\textwidth}
        \includegraphics[width=\textwidth]{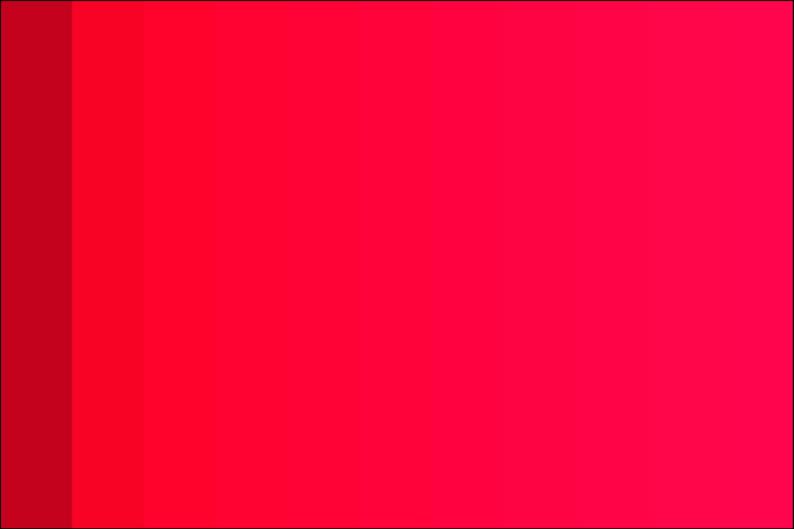}
    \end{subfigure}
    \hfill
    \begin{subfigure}[b]{0.193\textwidth}
        \includegraphics[width=\textwidth]{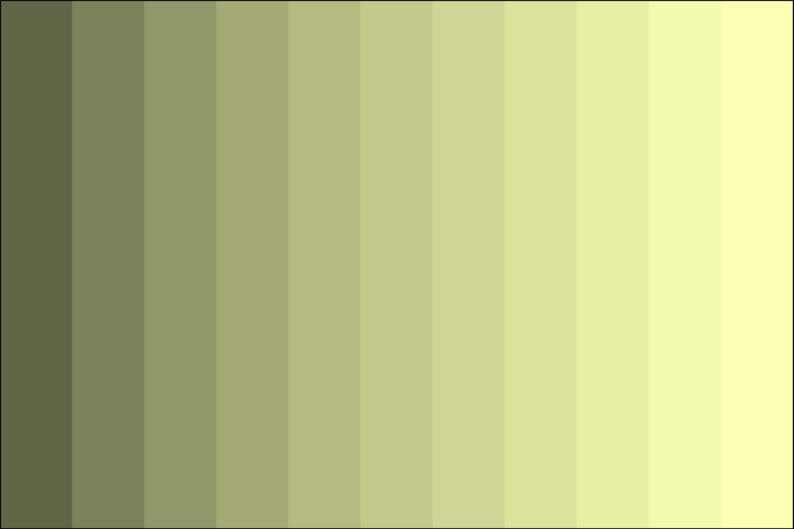}
    \end{subfigure}
    
    \vspace{0.2em}
    
    \begin{subfigure}[b]{0.193\textwidth}
        \includegraphics[width=\textwidth]{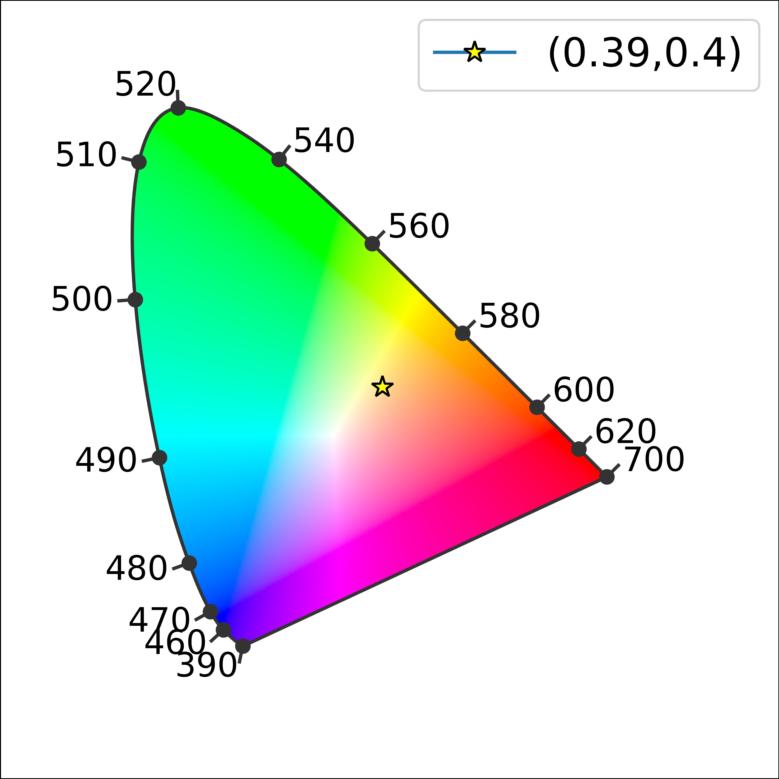}
        \caption{vehicle's front lights}
    \end{subfigure}
    \hfill
    \begin{subfigure}[b]{0.193\textwidth}
        \includegraphics[width=\textwidth]{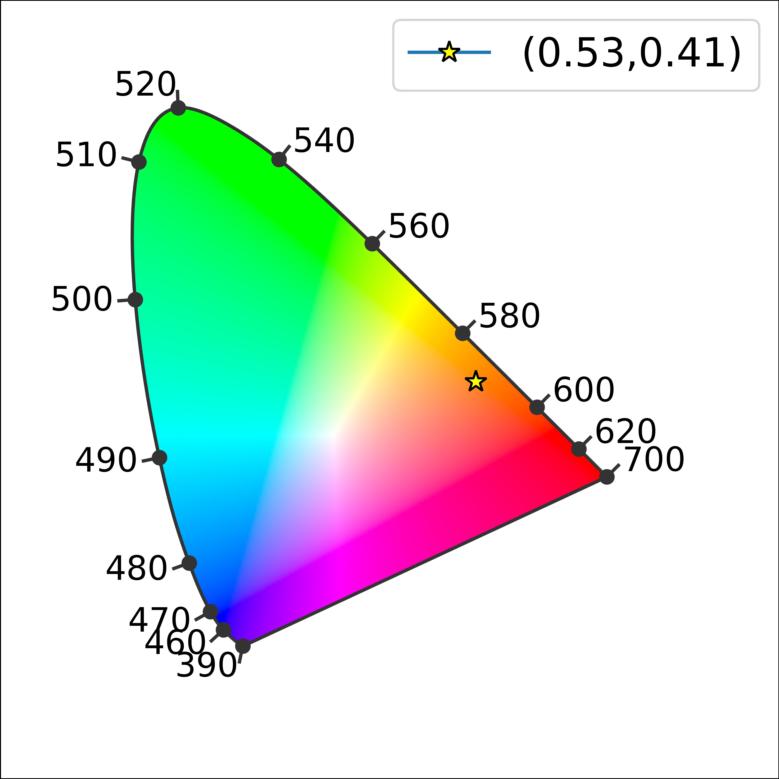}
        \caption{traffic light orange}
    \end{subfigure}
    \hfill
    \begin{subfigure}[b]{0.193\textwidth}
        \includegraphics[width=\textwidth]{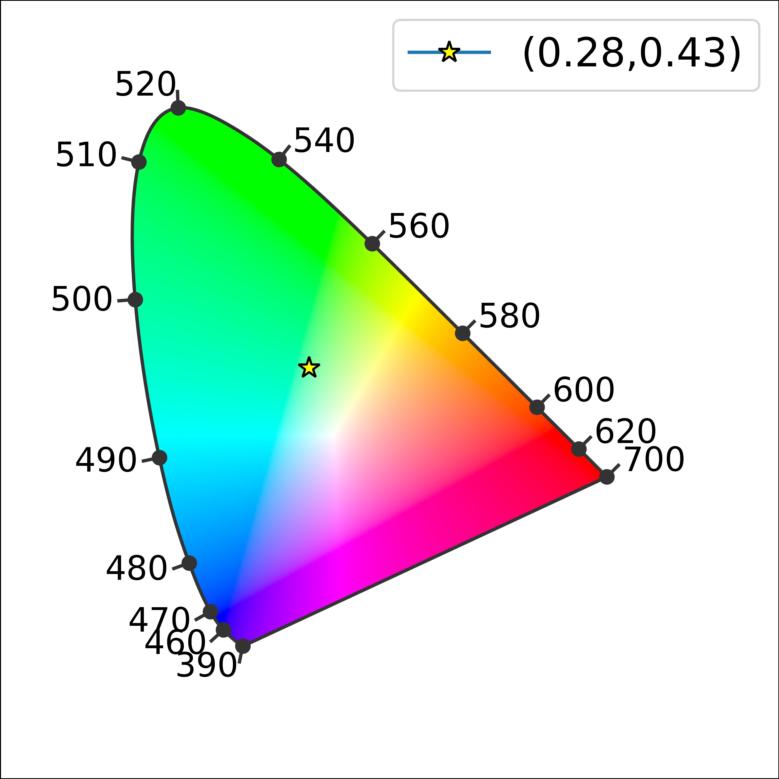}
        \caption{traffic light green}
    \end{subfigure}
    \hfill
    \begin{subfigure}[b]{0.193\textwidth}
        \includegraphics[width=\textwidth]{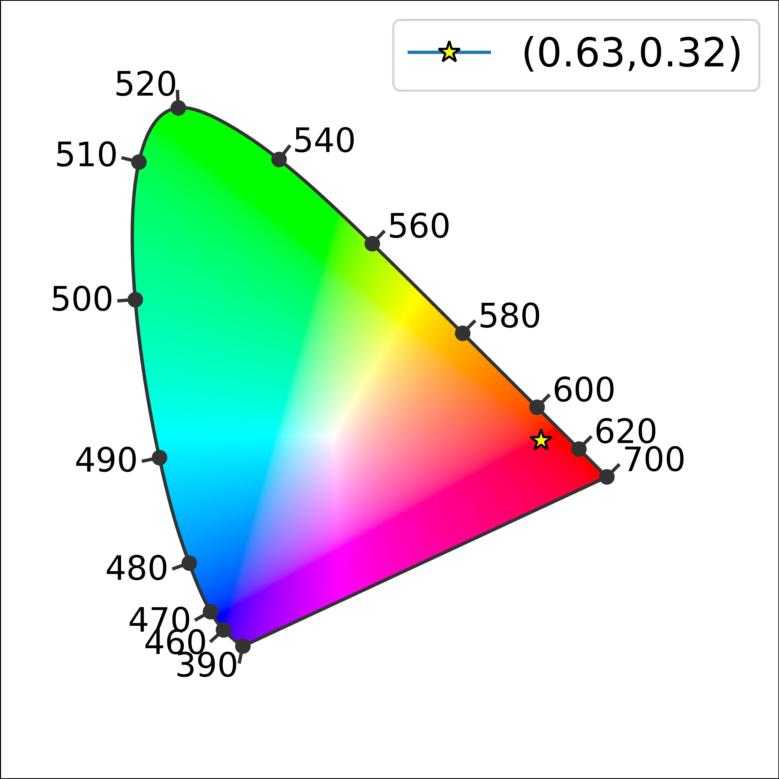}
        \caption{vehicle's rear lights}
    \end{subfigure}
    \hfill
    \begin{subfigure}[b]{0.193\textwidth}
        \includegraphics[width=\textwidth]{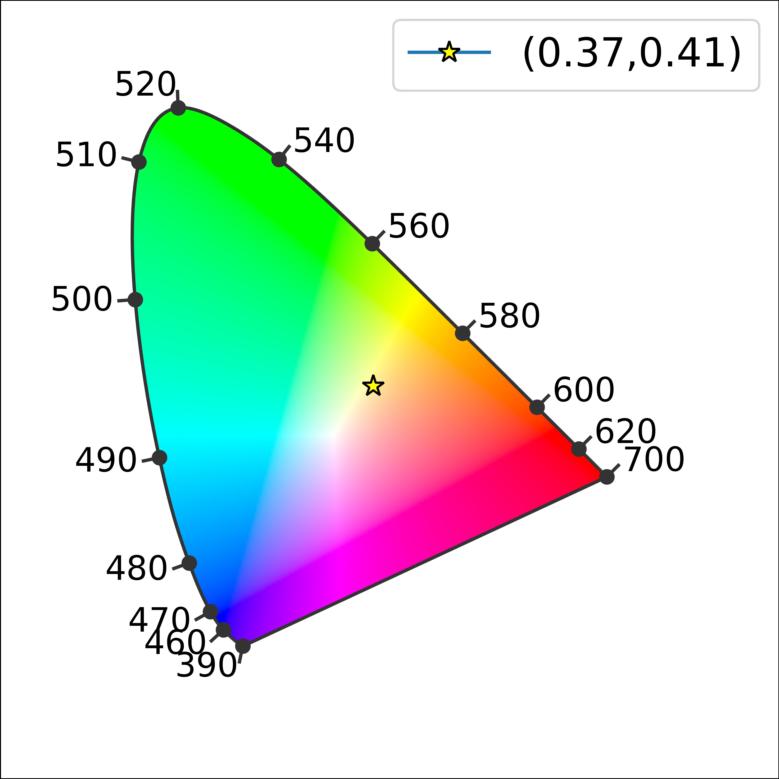}
        \caption{building's window light}
    \end{subfigure}
    
    \vspace{-0.2cm}
    
    \caption{\textbf{Nighttime Illuminants dataset samples.} In the first row, the gray card images from which the average chromaticity coordinates $(x,y)$ are calculated are illustrated. The generated color palette (varying luminance) of the sampled chromaticity coordinates is displayed in the second row. Lastly, the $(x,y)$ coordinates are plotted on the CIE 1931 $2^\circ$ Standard Observer chromaticity diagram in the third row.}
    \label{fig:sup:nighttime-illuminants}
    \vspace{-0.2cm}
\end{figure*}

\end{document}